\newif\ifieee
\ieeefalse

\ifieee
\documentclass[10pt,journal,compsoc]{IEEEtran}
\else
\documentclass[sort&compress,preprint]{myelsarticle}
\fi

\usepackage[final]{changes}

\usepackage{graphicx}                
\DeclareGraphicsExtensions{.pdf,.png,.jpg,.jpeg} 

\graphicspath{{figures/}{pictures/}{images/}{./}} 
\ifieee
\usepackage{cite}                      
\fi

\usepackage{hyperref}

\usepackage{microtype}                 
\PassOptionsToPackage{warn}{textcomp}  
\usepackage{textcomp}                  
\usepackage{mathptmx}                  
\usepackage{tabu}                      
\usepackage{booktabs}                  

\usepackage{xspace}
\newcommand{\sysname}{CBoxer\xspace}
\newcommand{\boxer}{Boxer\xspace}

\newcommand\trinaryview        {\emph{Trinary Distribution} view\xspace}
\newcommand\bandwidthassessview{\emph{Bandwidth Assessment} view\xspace}

\newcommand\reliabilitycurveview{\emph{Reliability Curve}   view\xspace}
\newcommand\perfconfview       {\emph{Performance Confidence} view\xspace}
\newcommand\uncertheatview     {\emph{Uncertainty Heatmap} view\xspace}
\newcommand\histogramview      {\emph{Histogram} view\xspace}
\newcommand\perfcurveview      {\emph{Performance Curves} view\xspace}
\newcommand\overallview        {\emph{Performance (Overall)} view\xspace}

\newcommand\rejectview         {\emph{Rejection Curve} view\xspace}
\newcommand\scatterview        {\emph{Scatterplot} view\xspace}
\newcommand\focusview          {\emph{Focus Item} view\xspace}

\newcommand\probcontrolpanel   {\emph{Probability Control}  panel\xspace}

\newcommand\selectpanel        {\emph{Selection Control} panel\xspace}

\newcommand\teaserfig{
  \centering
  \includegraphics[width=\linewidth]{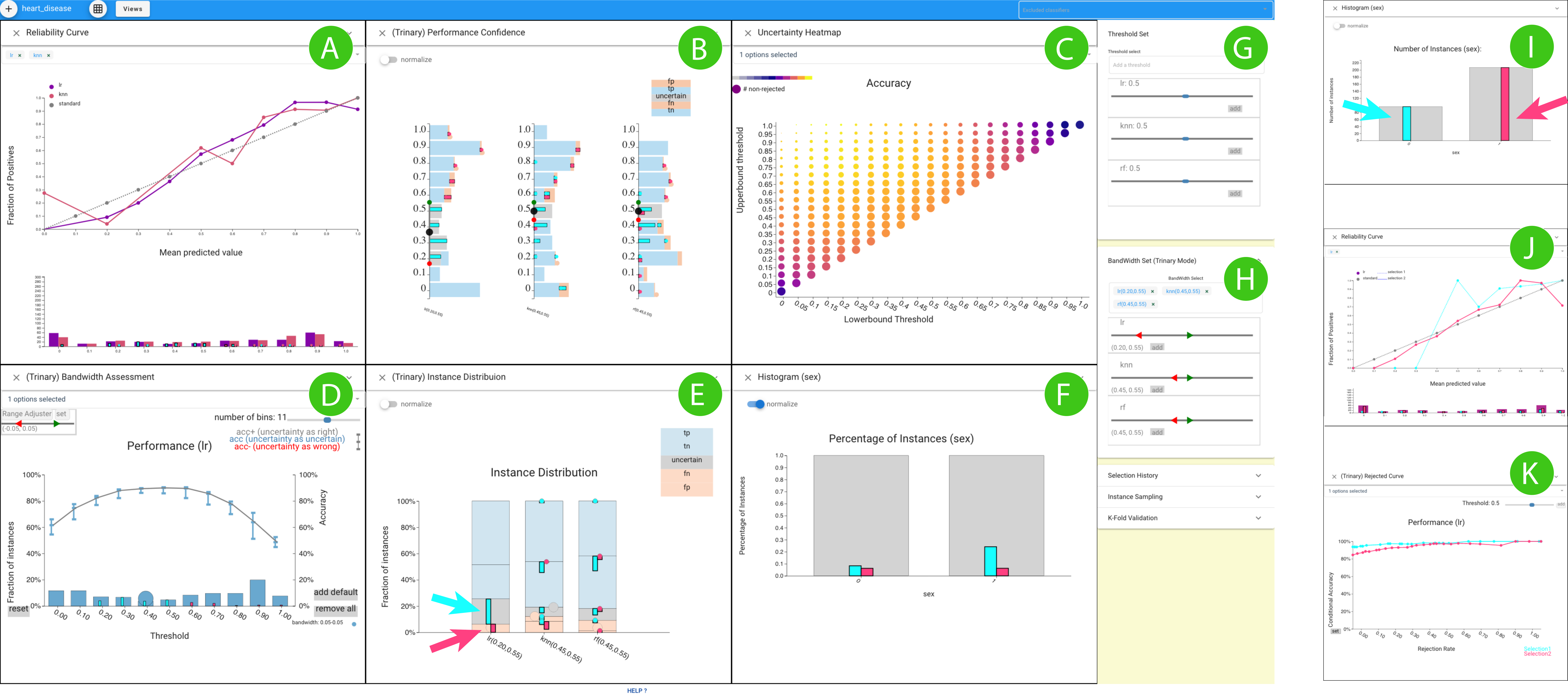}
  \caption{The CBoxer system assessing three classifiers for the disease prediction problem discussed in Section 1.1. The user has placed six views into the workspace: (A) \reliabilitycurveview, (B) \perfconfview, (C) \uncertheatview, (D) \bandwidthassessview, (E) \trinaryview, and (F) \histogramview. The \probcontrolpanel (G,H) is used to control thresholds.
  The user has selected the false positive (magenta) and uncertain (cyan) items for the LR classifier in (E) as indicated by the arrows (added to indicate click locations). These selections can be seen in other views, including (F) that shows that the classifier is more likely to be uncertain for men.
  The right columns shows a more detailed exploration. The sexes are selected in (I).
  The \reliabilitycurveview (J) shows how different groups provide differently calibrated classifiers, and the \rejectview (K) shows how these different calibrations provide different tradeoffs between rejection and performance.}
  \label{fig:teaser}\label{fig:heart}
}
\newcommand\boxerbasicsfig{
\begin{figure}[t]
  \centering\includegraphics[width=\columnwidth]{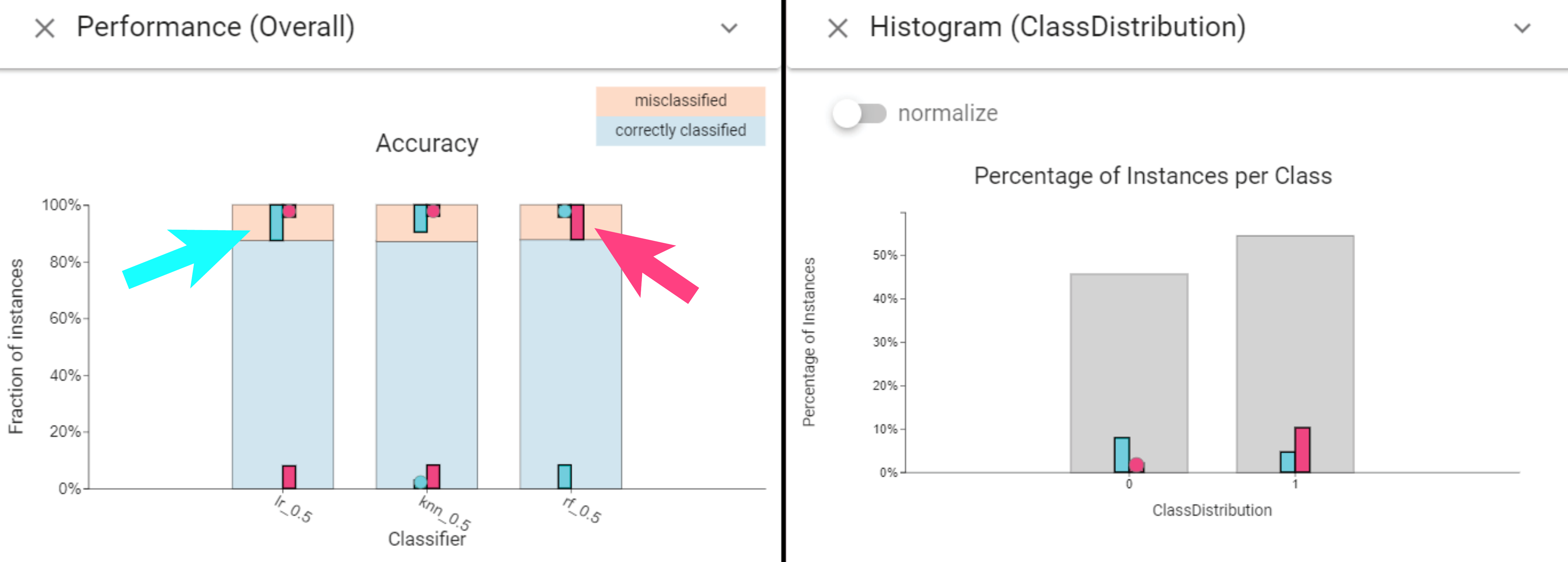}
  \caption{
    Basic views from Boxer \cite{boxer} allow for selections that are shown in other views. Bars in a bar chart or squares in confusion matrices can be clicked to set either of two selections. Left mouse click makes the first selection (cyan) and right click makes the second (magenta). These selections are shown across the interface.
    This example from \autoref{sec:example} shows the user selecting the errors from the LR and RF classifiers (cyan and magenta) to see how they are distributed differently across the classes.
    \label{fig:basicboxes}
    }
\end{figure}
}

\newcommand\bandassessfig{
\begin{figure}
  \centerline{
    \includegraphics[width=\columnwidth]{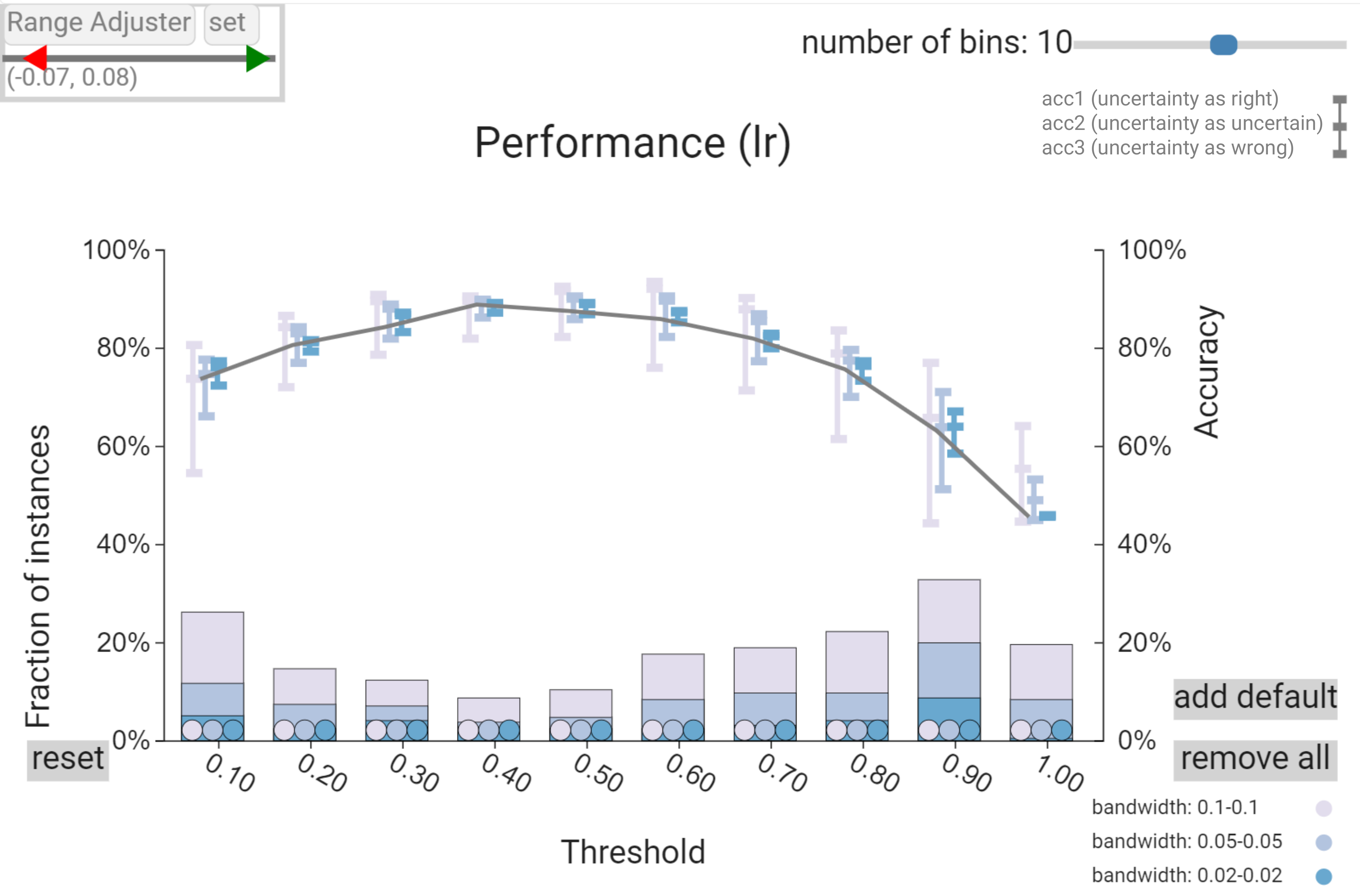}
  }
  \caption{
    The \bandwidthassessview summarizes the performance of a classifier over a range of thresholds and bandwidths.
    The line graph shows how a metric (accuracy shown) changes with threshold.
    Three bandwidths are selected for comparison, each is assigned a color. The error-bar-like glyphs show the range of metric possible with the different bandwidths depending on how one interprets rejected items. The top of the bar considers rejected items as correct, the bottom as incorrect. The bar chart at the bottom shows the number of rejected items
    (circles provide click targets for potentially small bars).
    \label{fig:bandassess}
  }
\end{figure}}

\newcommand\trinaryviewfig{
\begin{figure}
  \centerline{
	\includegraphics[width=\columnwidth]{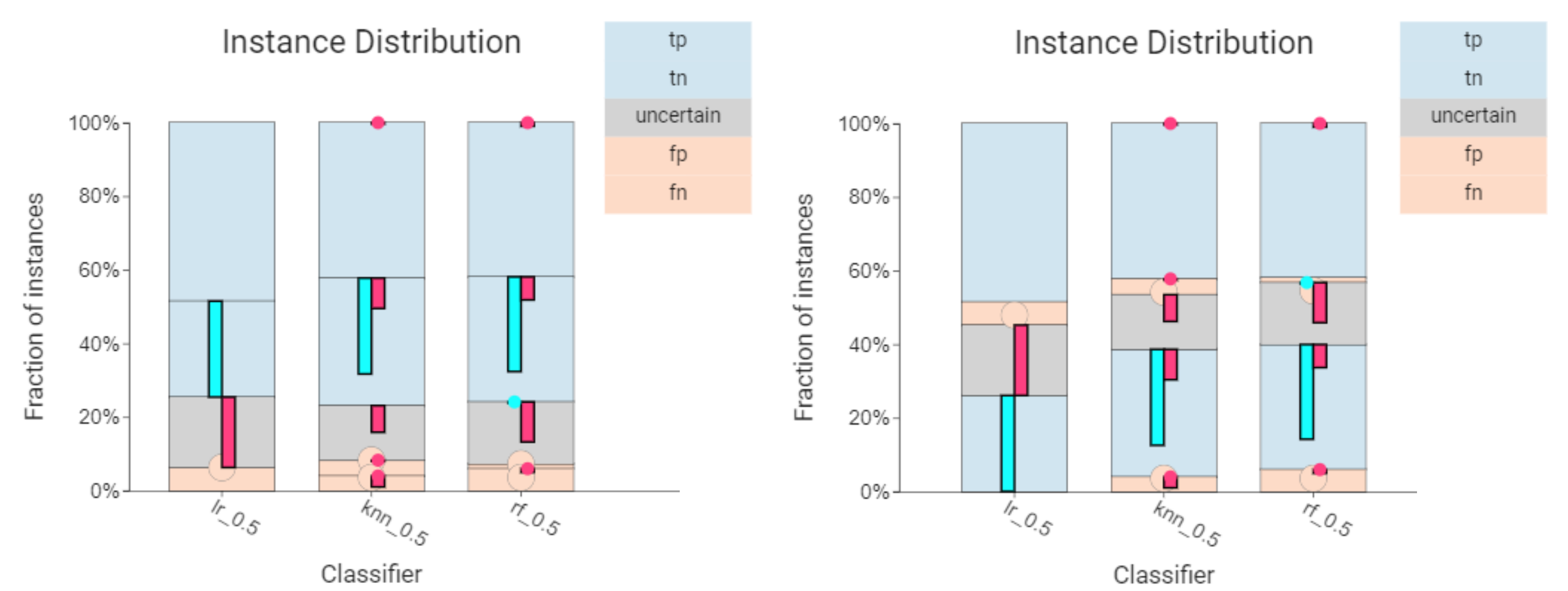}
  }
  \caption{
    The \trinaryview provides a summary of classification decisions for the classifiers. Different stack orderings allow for easier comparisons. Correctness mode (left) groups bars by correctness allowing for easier accuracy comparison, while score mode (right) places lower scoring groups lower in the bar. Dots are used to show small, but non-zero, values.
    The user has selected the true negatives (cyan) and uncertain (magenta) items in the LR classifier to see where those items appear elsewhere.
    \label{fig:trinary}
  }
\end{figure}
}

\newcommand\ucincome{
\begin{figure}
  \includegraphics[width=\columnwidth]{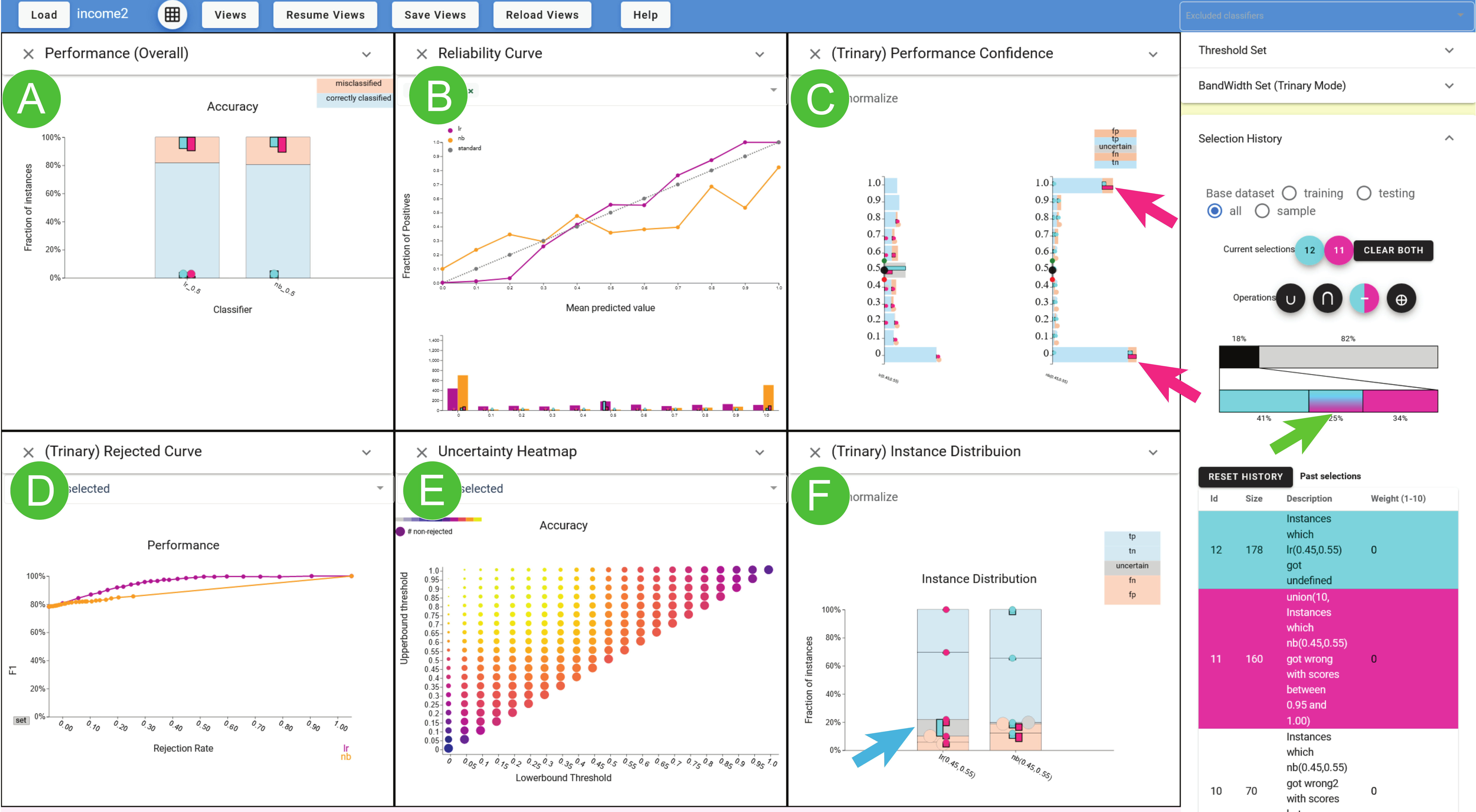} 
  \caption{     \label{fig:income-modelsel}
    Four views used in the model selection use case of \S\ref{sec:income-modelsel}.
    The (A) \overallview shows similar correctness, while the (B) \reliabilitycurveview shows differences in calibration. The (C) \perfconfview shows the differences in distributions. \rejectview (D) shows LR achives better F1 than NB when they have similar rejection rate.  \uncertheatview (E) also indicates the performance for NB is effectively unchanged across different threshold ranges.  (F) \trinaryview shows the classifications when items near the threshold (controlled by the \probcontrolpanel) are rejected.
    The arrows (drawn) show where the user clicked to select the items close to the threshold (rejected) by the LR classifier (cyan selection) and erroneous with extreme values for the nb classifier (magenta selection).
    The \selectpanel (E) shows the overlaps in the selections (green arrow).
  }
\end{figure}
}

\newcommand\cifarfig{
\begin{figure*}
  \centerline{
    \includegraphics[width=\linewidth]{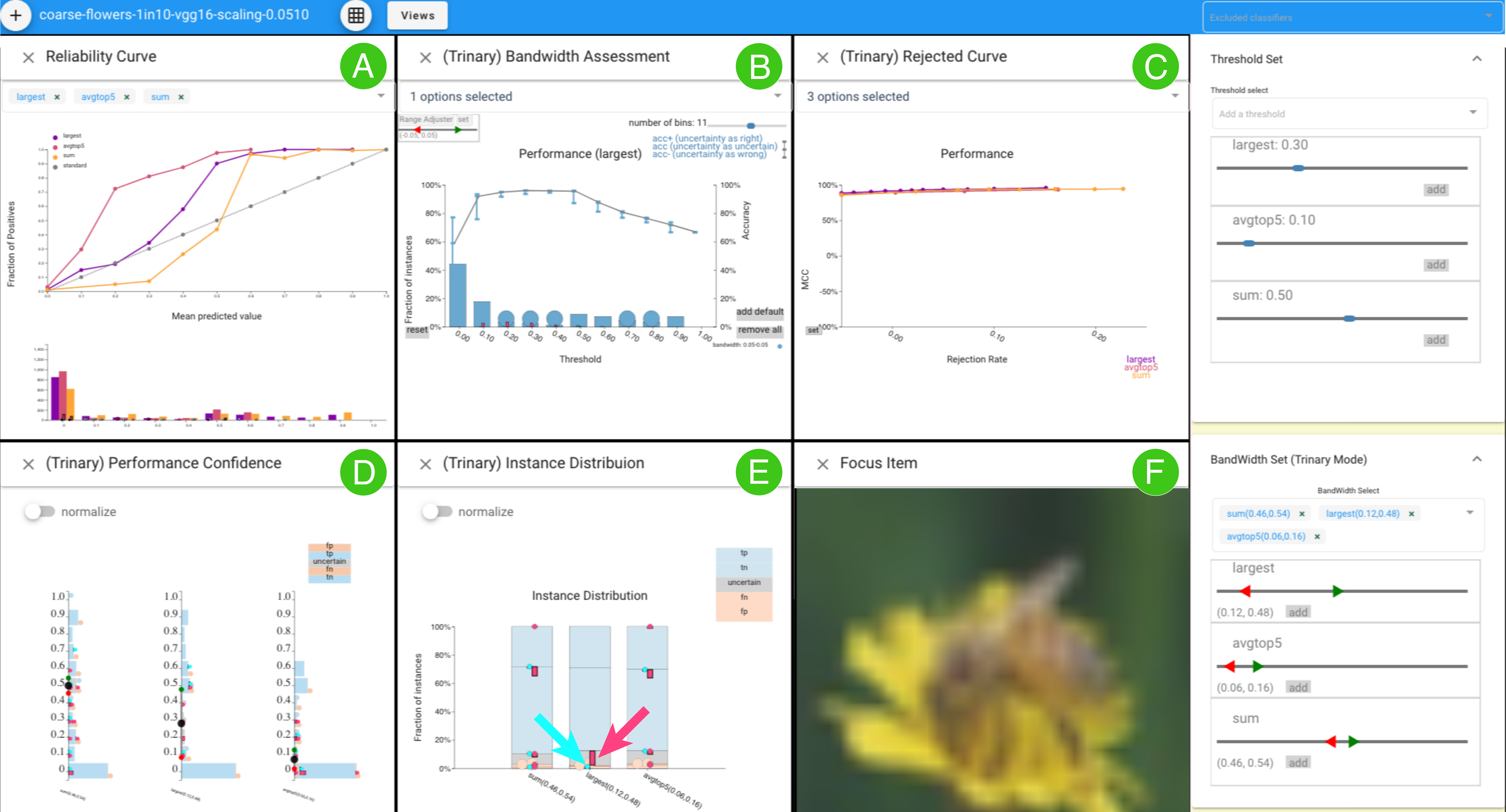}
  }
  \caption{
    \sysname in the flower image classification model selection use case of \S\ref{sec:cifar}.
    The \reliabilitycurveview (A) shows that each classifier achieves good performance, however they have quite different calibrations.
    The \bandwidthassessview (B - shown for one model) is used to find appropriate thresholds for each.
    The \rejectview (C) shows that all models have similar performance and provide good tradeoffs with rejection.
    The \perfconfview (D) allows detailed analysis and selection.
    The \trinaryview (E) allows selection of rejected and incorrect items; while the models make the same number of errors and rejections, they are different.
    The \focusview (F) allows scanning through the selected errors to identify patterns. Here, the image contains a flower and a bee and is, therefore, labeled insect.
    \label{fig:cifar}
  }
\end{figure*}}

\newcommand\ucwinefig{
\begin{figure}
  \centerline{\includegraphics[width=\columnwidth]{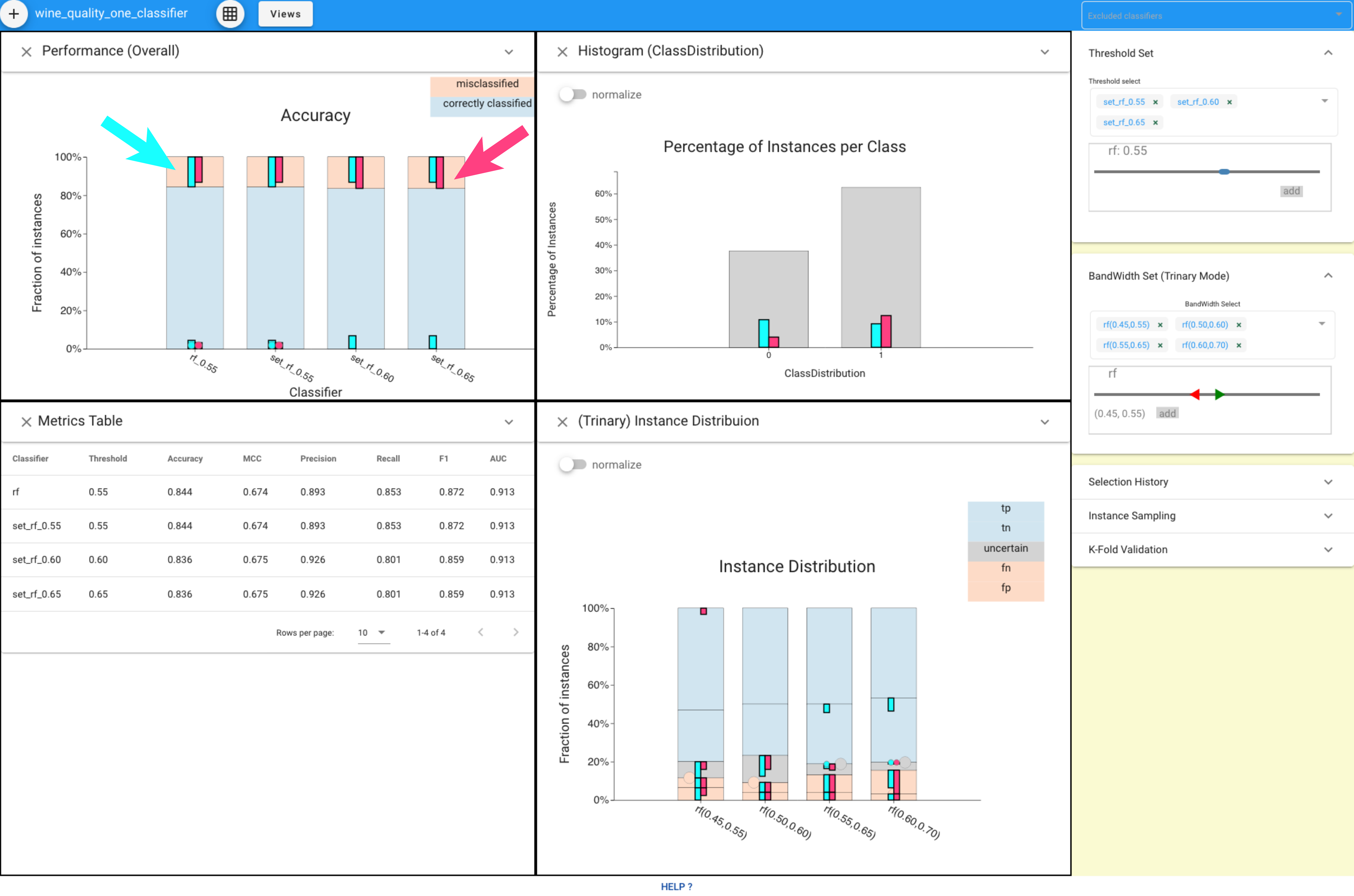}}
  \caption{
    Views used in the wine threshold use case \S\ref{sec:wine-threshold}
    \label{uc2-wine-threshold}. Variants of the classifier created for different thresholds have similar accuracies and other metrics. Selecting the errors in the lowest threshold (cyan) and highest (magenta) shows that there are differences.
    \label{fig:uc2-wine-threshold}
  }
\end{figure}
}
\newcommand\ucdatesperf{
\begin{figure}
	\centerline{\includegraphics[width=\columnwidth]{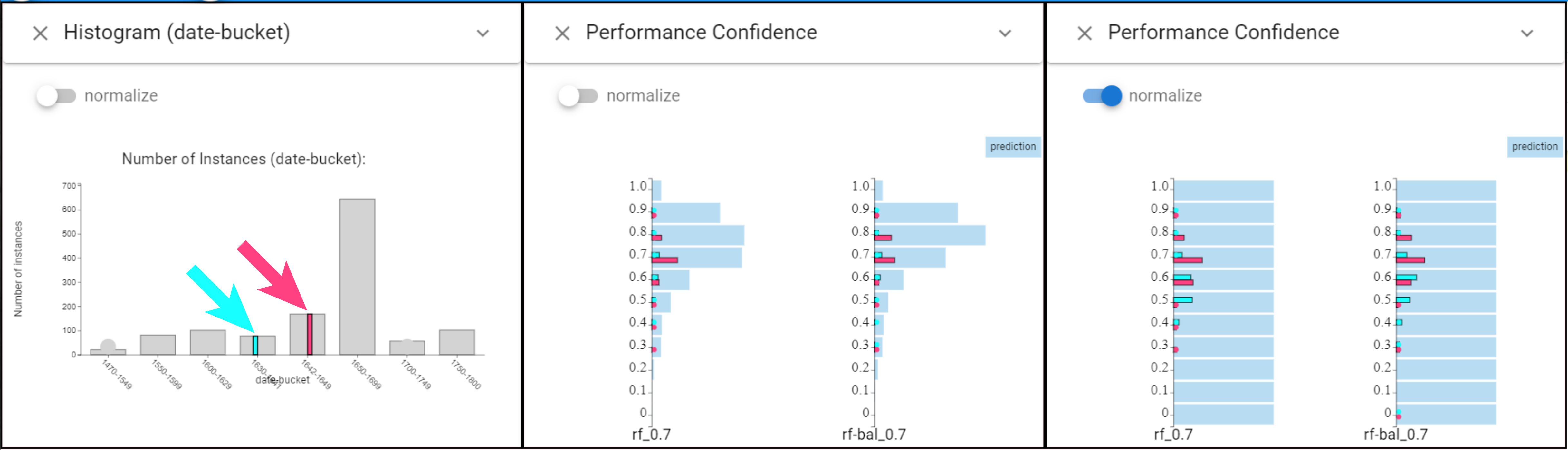}}
	\caption{
		CBoxer applied to the English literature date classification scenario.
		The user has selected the time periods before an after the critical date as the selections in the \histogramview of dates (left). The \perfconfview (shown with and without normalization, but without correctness coloring) show that these groups tend to have scores near the threshold.
		\label{fig:lit-dates}
	}
\end{figure}
}
\newcommand\ucdatesfiglengths{
\begin{figure}
  \centerline{\includegraphics[width=\columnwidth]{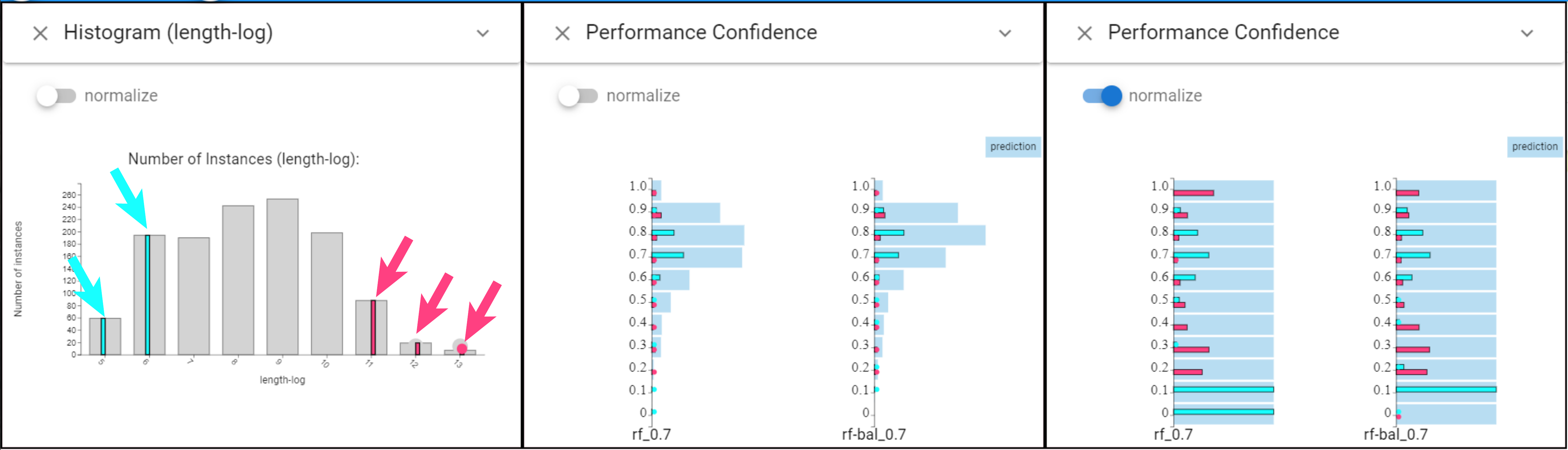}}
  \caption{
    CBoxer applied to the English literature date classification scenario.
    The user has selected the shortest (cyan) and longest (magenta) documents in the \histogramview. The \perfconfview (shown with and without normalization, but without correctness coloring) show that long documents are generally father from the threshold. The large bars at the bottom of the normalized \perfconfview are outliers with a single document in the bin.
    \label{fig:lit-lengths}
  }
\end{figure}
}

\newcommand\scatterfig{
\begin{figure}
  \centerline{
    \includegraphics[width=\columnwidth/3]{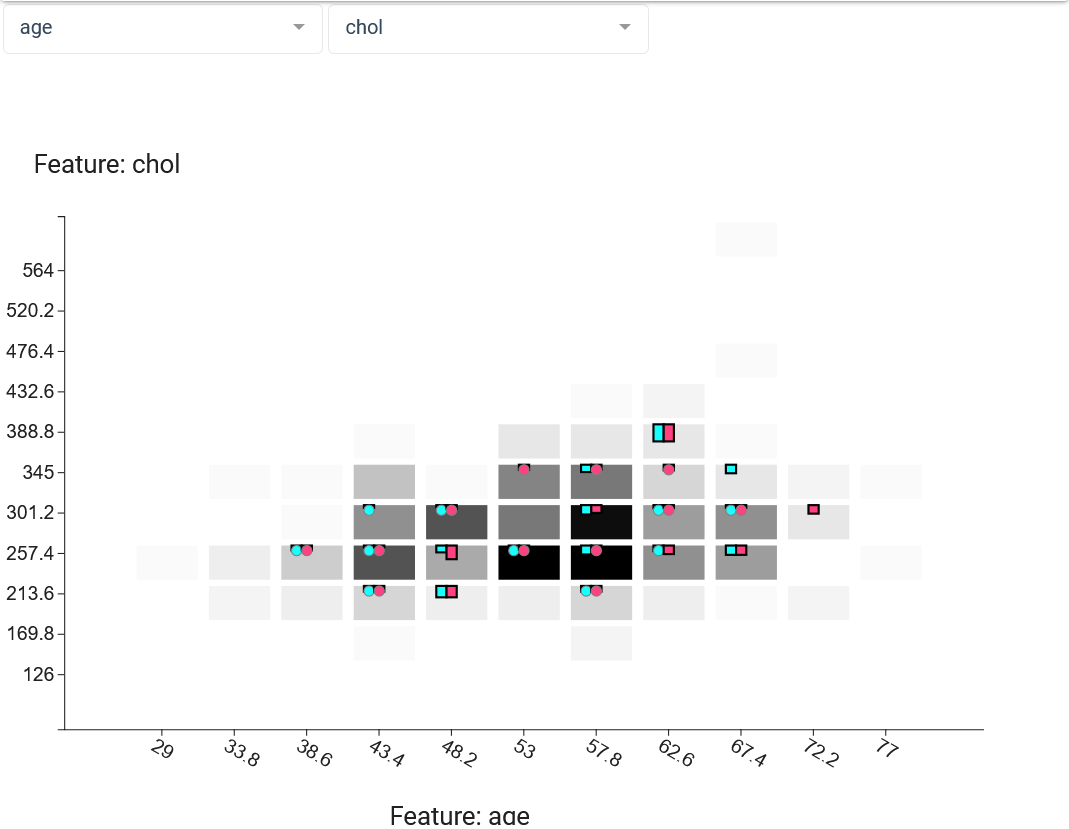}
    \includegraphics[width=\columnwidth/3]{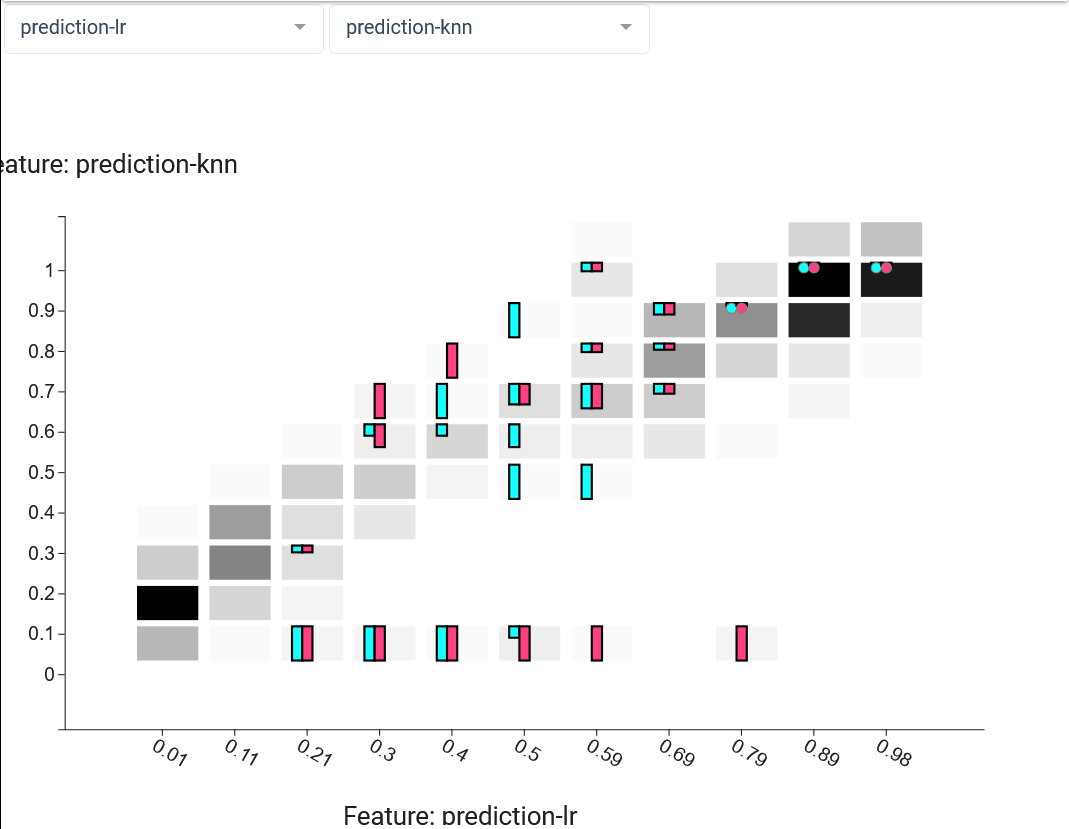}
    \includegraphics[width=\columnwidth/3]{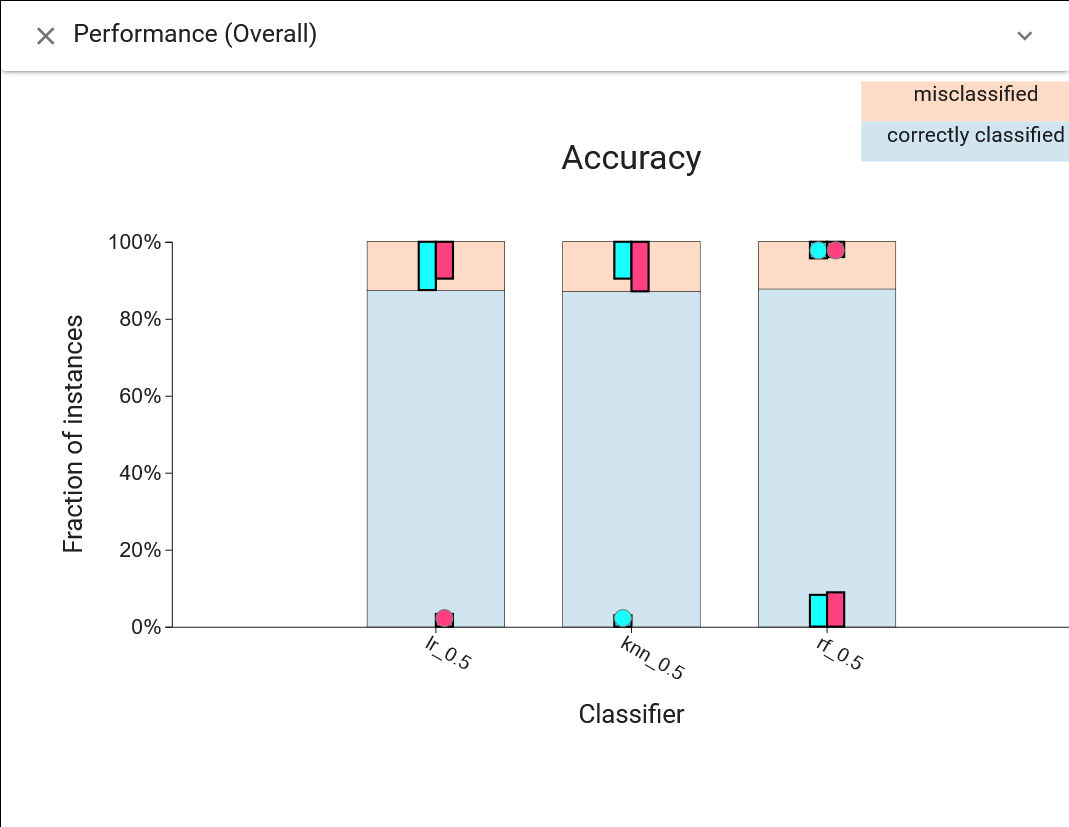}
  }
  \caption{
  \label{fig:scatter}
  \scatterview views allow the user to select any parameters. The data is from the example of \autoref{sec:example}, \autoref{fig:teaser}. (left) Two features (age and cholesterol) to look for patterns. (center) The scores from the LR and KNN classifiers are shown to assess their correlation. For both, the \overallview (right) is used to select the errors of the LR classifier as cyan and the KNN classifier as pink.
  }
\end{figure}
}

\begin{document}

\title{Trinary Tools for Continuously Valued Binary Classifiers}

\ifieee
\else
    \author{Michael Gleicher}
        \ead[url]{gleicher.sites.cs.wisc.edu}
        \ead{gleicher@cs.wisc.edu}
    \author{Xinyi Yu}
    \author{Yuheng Chen}
    \address{Department of Computer Sciences}
    \address{University of Wisconsin -- Madison, Madison, WI, USA}
\fi


\ifieee
\else
    \begin{abstract}
    Classification methods for binary (yes/no) tasks often produce a continuously valued score.
Machine learning practitioners must perform model selection, calibration, discretization, performance assessment, tuning, and fairness assessment.
Such tasks involve examining classifier results, typically using summary statistics and manual examination of details.
In this paper, we provide an interactive visualization approach to support such continuously-valued classifier examination tasks.
Our approach addresses the three phases of these tasks: calibration, operating point selection, and examination.
We enhance standard views and introduce task-specific views so that they can be integrated into a multi-view coordination (MVC) system.
We build on an existing comparison-based approach, extending it to continuous classifiers by treating the continuous values as trinary (positive, unsure, negative) even if the classifier will not ultimately use the 3-way classification. We provide use cases that demonstrate how our approach enables machine learning practitioners to accomplish key tasks.

    \end{abstract}
\fi

\maketitle



\begin{centering}
  \fbox{\parbox{4in}
  {\textcolor{red}{
  Author's version. This paper is accepted to appear in the journal \emph{Visual Informatics}. The official version is available Open Access at \url{https://doi.org/10.1016/j.visinf.2022.04.002}.
  }}}
\end{centering}

\ifieee
\else
    \section{Introduction}
\fi

Machine learning practitioners examine the results of models as part of developing and deploying classifiers.
For classifiers, examining results over a testing set is part of model selection (i.e., choosing the most appropriate classifier and tuning its hyper-parameters appropriately) but also in assessing whether a classifier is good enough (Can it be trusted?) and fair (Does it treat subgroups equivalently?).
Practitioners examine results in aggregate, looking at summary statistics or graphs of performance over testing sets, and drill into these results to identify interesting subgroups or specific items as part of understanding performance.
Examination is often performed manually using scripting within the development workflow.
Recent interactive visualization approaches have been introduced to help with the assessment of classifier results.

We consider binary classifiers where models compute continuously valued scores.
These scores are often thresholded to provide a binary decision. In some cases, the score is also used for \emph{rejection} \cite{chowOptimumRecognitionError1970,landgrebeInteractionClassificationReject2006} where the classifier chooses not to make a decision. We term such classifiers as \emph{trinary} as they provide one of three results (positive, unsure, negative). \added{Assessing classifier scores involves three phases of tasks: \emph{calibration,} where the scores are examined to assess how well they correspond to possible decisions (e.g., do the outputs meaningfully quantify  confidence or uncertainty?); \emph{operating point selection,} where the thresholds for decisions and (optionally) rejection are determined; and \emph{examination} where the decisions made by the classifier are explored in detail.}
These tasks often require human involvement as the decisions require judgment of risks and rewards.
However, summary statistics of continuous classifiers (e.g., P/R or ROC curves or area-under-curve metrics) do not directly connect to tasks and no standards have emerged to assess them \cite{condessaPerformanceMeasuresClassification2017}.
Similarly, interactive tools have not considered the full range of tasks or connecting detailed examination to the broader statistics.

In this paper, we introduce an interactive visualization approach that aids practitioners with result examination tasks for continuously-valued binary classifiers.
\added{We provide a system that allows a user to select and combine a variety views.
The views build on existing tools: we extend summary statistics and charts with enhanced designs and new visualizations. These are designed to address specific tasks and afford coordinated interactions that enable flexible uses.}
\deleted{We build on the existing tools: summary statistics and charts for continuous models and interactive approaches for model result exploration. }
\added{We integrate these views with support for comparison.}
\deleted{We extend the standard statistics and charts with interactive elements that enable them to work together in a multi-view application, we provide new views that address specific tasks, and enhance an existing comparison-based approach to enable its use with continuous models.}
Comparison enables understanding the differences between models or hyper-parameter settings, or between different groups of items within a model.
Because these tools are used to estimate hyper-parameters (thresholds), our approach integrates sampling to allow separate sets for estimation and validation.

Our target audience is machine learning practitioners, ranging from students to experienced developers and researchers. This audience is familiar with the standard terminology, metrics, and charts. From a design perspective, this means an approach must include the common views that such users demand for basic tasks, but extend them to integrate into workflows that support tasks that are inconvenient to do with standard script-based tooling. For example, users typically create standard curves (ROC or P/R) to assess a classifier's correctness and estimate good threshold values, but they may need augmentations to these standard views to help identify relevant examples and new views that can help them evaluate rejection criteria.
\added{Our work shows how we can integrate the measures used in practice into interactive views that assist in interpreting the results, but also connecting them to the underlying examples through interaction.}

We focus on examining classifier results: we treat the classifiers as black boxes and consider only input/output pairs.
\added{A key idea is that a classsifier's scores are not always meaningful: an approach that uses them effectively must consider \emph{calibration} to examine that the scores have meaning as a component of their use.}
A second key idea is that by integrating support for trinary (yes/unsure/no) decision making, we can support specific tasks for rejection classifiers, but also use this trinary classification as a strategy to organize analysis where the ultimate classifier will not use rejection.

Our core contribution is an approach to assessment of continuously-valued binary classifiers that is sufficiently flexible to adapt to a wide range of tasks, yet allows for detail oriented analysis.
We show how the variety of summaries used with such classifiers can be adapted to work within an interactive tool, and introduce new views to assist with tasks that are not well-served by existing tools.
We show how these views can be coordinated to serve detailed exploration and comparison tasks by extending the select-and-compare approach introduced in prior systems.
The idea of trinary classification enables us to connect continuously \replaced{valued}{values} classifiers with discrete-group-based tools.
Our approach, including these novel techniques, has been integrated into the \sysname\footnote{\sysname is not a separate system from the original Boxer. It is a new version that adds support for continuously valued classifiers. The system is available at \url{http://pages.graphics.cs.wisc.edu/Boxer}.} classifier comparison and analysis system, which provides an open source prototype of the ideas.

\begin{figure*}[t]
  \teaserfig 
\end{figure*}
\subsection{An Example}
  \label{sec:example}
  \label{sec:intro-example}
We begin with an example to introduce the key concepts in our approach before giving the details of the views.
We consider the tasks of model selection, calibration, and fairness assessment.
This example uses the Heart Disease Prediction dataset, a standard data set used in machine learning education \cite{uci}. Classifiers are trained to predict if a patient will develop a disease (binary decision).
The types of errors matter: a false negative (not warning of a potential problem) may be worse than a false positive (which may cause fear).
A ``no-prediction'' option is viable, recommending that patient is tested further rather than making a potential error.

We use our \sysname system to compare 3 classifiers trained with different methods. Three classifiers are trained with different methods (logistic regression (LR), k-nearest neighbor (KNN) and random forest (RF)).
These are evaluated on a set of testing data.
The results are loaded in \sysname (\autoref{fig:heart}).

\sysname can show standard summary metrics as well as allowing for detailed examination. For example, while models have similar accuracy (87\%) and F1 scores (.88), RF has substantially worse recall which is important given the cost of false negatives.
\sysname allows for examining a variety of metrics or the confusion matrices to perform model selection.
Alternatively, \sysname's comparative mechanisms allow subsets of the items to be selected and compared to examine \emph{where} errors occur.
\autoref{fig:basicboxes} demonstrates how selection of errors from LR and RF enables comparing them to see that RF's errors are mainly patients with disease.
Examining the rejection curve shows that performance still lags even after tuning.
\sysname enables model selection by bringing together multiple methods for comparing metrics in an integrated way.

We select between LR and KNN by considering the utility of their scores. We seek a classifier that is \emph{calibrated} such that its scores
useful information about the predictions,
enabling rejection if a decision lacks confidence.
While a summary measure of score quality (AUC) suggests the classifiers are similar, \sysname allows us to examine the difference in detail. It provides specialized views, discussed in \autoref{sec:tuning} and \autoref{sec:compare-views}, to understand the distributions of the scores and how they relate to correctness. The \reliabilitycurveview (Fig. 1A) shows that both are reasonably well calibrated, while KNN has more errors with very low scores. The \perfconfview (Fig. 1B) also shows this: KNN has many false negatives where it provides extremely low scores. In \sysname, we could select these instances for further examination.
These confident false negatives are problematic for the scenario.
\sysname integrates aggregate and detail selection approaches allowing for specialized analysis.

We select the LR classifier. Because the LR classifier is well calibrated, we can use its score to \emph{reject} predictions that are unlikely to be correct: we create a \emph{trinary} classifier where middling scores are rejected. This requires defining upper and lower thresholds. Our approach supports for exploring the trade-offs between rejection and correctness (\autoref{sec:tuning}). In \sysname, users can manipulate thresholds to see their impacts across various performance summary views.
We also provide a number of views to examine trade-offs. The  \bandwidthassessview (Fig. 1C) and \uncertheatview (Fig. 1D) provide summaries about how the two thresholds work together to provide different tradeoffs, while more traditional views aid in considering the parameters independently.
\sysname provides support for exploring the tradeoffs in trinary classifiers enabling more flexible tuning.

We choose settings that provide high recall with a reasonable (20\%) rejection rate.
We can view the proportion of classified instances in the \trinaryview (Fig. 1E).
We investigate the fairness of this classifiers by choosing its errors (pink selection in Fig. 1E) and seeing that they are evenly distributed across sexes, (Fig. 1F). This meets the typical fairness criterion used by systems such as \cite{Cabrera2020,Ahn2020}). However, selecting the rejected items (cyan selection in Fig. 1E) we see that men are far more likely to receive uncertain predictions.
We can examine this fairness issue in detail (Fig 1. I,J,K). We select the sexes as groups and compare them using the \reliabilitycurveview (Fig. 1J) and \rejectview (Fig. 1K).
\sysname enables detailed exploration of classifier results, allowing for identifying groups based on continuous or trinary properties and examining their impact in various comparisons.

\section{Background and Related Work}
  \label{sec:related}
  \label{sec:background}

\noindent\emph{Performance Assessment for Binary Classifiers:}
Classification models are assessed empirically: a practitioner considers a testing set with known correct answers.
While the most basic assessment counts the proportion correct (accuracy), other metrics consider the kinds of errors and/or the likelihood of guessing \cite{provostCaseAccuracyEstimation1998}. A variety of metrics summarize performance in different ways, see \cite{Powers2011} for an overview. For example, recall emphasizes false negatives while the Matthews Correlation Coefficient (MCC) considers the base rates to compare a classifier to chance \cite{parkerAnalysisPerformanceMeasures2011,chiccoAdvantagesMatthewsCorrelation2020}.
These metrics summarize 
performance into a single number, hiding the mixture of kinds of errors and details of which items the classifier gets right or wrong.

\noindent\emph{Assessment of Continuously Valued Classifiers:}
Many binary classifiers provide a continuously valued score
that is thresholded.
Adjusting the threshold allows for trade-offs in the types of errors.
The appropriate balance depends on the scenario.
For a given \emph{operating point} (i.e., value of the threshold(s)), binary assessment metrics can be applied.
Metrics can be optimized, but this requires knowing criteria and weights appropriate for the scenario.
Classifiers can be compared at specific operating points.  Non-threshold metrics \cite{lingAUCBetterMeasure2003} allow for more general comparison of models. Such metrics include the Receiver Operating Characteristic (ROC) curve and the Precision/Recall curve.
The ROC is often summarized by the area under the curve (AUC).
Such summary metrics hide details about the errors.

\noindent\emph{Calibration:}
Use of the classification score beyond thresholding assumes that the value of the score has meaning, i.e., that more extreme scores are more likely or confidently correct.
Early approaches to understanding the distribution of prediction scores provided methods for assessing calibration \cite{brierVerificationForecastsExpressed1950}, and a rich set of metrics are available \cite{gneitingProbabilisticForecastsCalibration2007}.
Simple visualizations are common to summarize calibration, and inspire our \reliabilitycurveview and \perfconfview.
\emph{Our work enhances the common views to integrate into workflows that support detailed assessment.}

Not all classification models are able to provide useful scores \cite{schwarz2019guess,guoCalibrationModernNeural2017}.
Calibration is a desirable property for a classifier \cite{Kuleshov2015_CalibratedStructedPrediction,zhongAccurateProbabilityCalibration2013,guoCalibrationModernNeural2017}. An \emph{over-confident} classifier (that gives extreme scores in situations where is is incorrect) or \emph{under-confident} classifier (that gives scores near the threshold even when correct) means that the scores cannot be used. In contrast, a \emph{calibrated}
classifier whose scores reflect the likelihood of its correctness allows the score to be used in interpreting its results\cite{brierVerificationForecastsExpressed1950,gneitingProbabilisticForecastsCalibration2007}.

The desire to have calibrated scores has lead to a long-standing effort in the machine learning community to extend classification methods (e.g., \cite{plattProbabilisticOutputsSupport1999,lingBayesianClassifiersAccurate2002,zhongAccurateProbabilityCalibration2013,Kuleshov2015_CalibratedStructedPrediction}). Other methods post-process classifier outputs to improve calibration \cite{guoCalibrationModernNeural2017}. Another approach is to create a separate score of the reliability of a classifier as a trust score \cite{jiangTrustNotTrust2018} or rejection score \cite{cortesLearningRejection2016}. Such scores can be combined with classifier predictions in our approach.
Some views including calibration curves and reliability diagrams ~\cite{predictingGoodProbabilities2005,scikit-learn,leathart2020temporal} have been developed for assessing calibration. We integrate these into our approach to support detailed assessment of classifier calibration.

\noindent\emph{Trinary Classification:}
\label{sec:trinary}
In some scenarios, classifiers are given the option to \emph{reject}, choosing not to make a decision if the likelihood of error is too high \cite{chowOptimumRecognitionError1970}. The rejection criteria may be \emph{embedded} into the design of the classifier, or may be performed \emph{post-hoc} based on the classifier's decision scores if they are sufficiently calibrated. We focus on the latter case.

Rejection classifiers have two thresholds that provide trade-offs between error types and failing to predict. With an embedded classifier, these may be different thresholds for the score and rejection level; for post-hoc rejection they may be viewed as a ``bandwidth'' around a single central threshold or as an upper and lower bound. The selection of an operating point requires choosing these two parameters. Given costs of the error types and a model of the posterior likelihoods of the population, optimal parameter values can be determined \cite{chowOptimumRecognitionError1970}. However, in practice, the costs may be hard to determine and vary, and performance can only be assessed over testing populations. Therefore, practitioners typically need to assess the trade-offs carefully.


Performance metrics for rejection classifiers must consider both correctness and number of rejections. According to \cite{condessaPerformanceMeasuresClassification2017}, ``There is no adequate standard measure for the assessment of the performance of a classifier with rejection.'' Single numbers fail to capture the tradeoffs between correctness and rejection rate.
Existing metrics (e.g., \cite{condessaPerformanceMeasuresClassification2017}) do not consider error types or class imbalances.

Rejection classifiers are typically examined using curves that show their correctness/rejection trade-offs\cite{hanczarPerformanceVisualizationSpaces2019}.
While a number of options exist \cite{hanczarPerformanceVisualizationSpaces2019}, the most common approach is to plot correctness (accuracy or error rate) as a function of rejection rate, known as an Accuracy Rejection Curve (ARC) \cite{nadeemAccuracyRejectionCurvesARCs2009}.
ARCs (and their variants) show the impact of different rejection rates, and, therefore, the rejection thresholds or bandwidths that determine them. However, they do not allow for considering both thresholds in a coordinated fashion. \cite{landgrebeInteractionClassificationReject2006} suggested the use of a 3D version of ROC curves, but such curves are challenging to interpret.

ARCs are useful not only for determining rejection thresholds, but also for comparing the calibration of classifiers \cite{nadeemAccuracyRejectionCurvesARCs2009}. This is used in selecting classifiers that provide a meaningful range of trade-offs. However, the approach has not been coupled with other tools for detailed analysis, and does not support the tuning of both rejection rate and decision threshold.

\subsection{Interactive Tools for Classifier Assessment}
Interactive tools for understanding machine learning models are motivated by many reasons, see Gleicher~\cite{Gleicher2016} or Lipton~\cite{Lipton2016} for surveys.
Our work falls into the broad category of \emph{black box} methods that focus on examining results of trained models, in contrast to methods for creating transparent models, providing explanations of decision, or inspecting model internals.

\label{sec:related-tools}
\label{sec:tools-for-specific}
Prior tools for interactive examination of the outputs of continuously valued classifiers are similar to our work.
Like our approach, ModelTracker~\cite{amershi2015modeltracker} and Prospector \cite{Krause2016} consider  assessment of binary classifier output, providing summary views to identify details and set thresholds. Their designs influenced ours. ModelTracker focuses on specific tasks, e.g., adjusting the threshold and identifying instances. \cite{ren2017squares} and \cite{alsallakh2014visual} provide related approaches for multi-class probabilities. These tools focus on multi-class issues, such as cross-class confusion, rather than single class thresholding. \emph{None of these approaches consider rejection, comparisons among subgroups, or supporting sampling to mitigate generalization mistakes.}

A number of tools have focused on specific scenarios for black-box classifier result analysis.
Some tools, such as \cite{heyen2020clavis}, focus on aggregate summaries over large collections of classifiers.
Others, such as \cite{zhang2018manifold}, focus on identifying relationships between outputs and features.
Other tools \cite{Cabrera2020,Ahn2020} focus on identifying subgroup fairness issues.
\cite{Wexler2020whatif} and \cite{boxer} describe systems that enable identifying subgroups for comparisons. \emph{Our work extends these ideas to the needs of continuously valued binary classifier assessment.}

Threshold tuning is often automated, with a historic tradition of optimization \cite{chowOptimumRecognitionError1970}. However, specific scenarios require trade-offs to be encoded as costs, and for good models of the likelihood of errors to exist. Tools, such as \cite{das2020questo,Kapoor2010}, have considered interactive refinement of the objectives used for parameter tuning in classifier construction.
\emph{Our approach provides tools that allows users to explore parameters to find tradeoffs suitable for their scenarios.}

\subsection{Detailed Assessment of Classifier Results}
\label{sec:discrete-tasks}
Detailed exploration of classifier results beyond summary statistics identifies groups or individual items and connects them to performance. Traditionally, practitioners use scripting in their development environments. However, recent tools support detailed examination (\autoref{sec:related-tools}). A range of assessment tasks benefit from detailed exploration. Tasks of \emph{model selection} benefit from considering if performance problems exist on more or less important items, \emph{model tuning} benefits from identifying items where problems occur so they can be addressed, \emph{data checking} identifies problems in the data using classifier results, and \emph{subset fairness} considers whether performance is balanced across latent groups.
Methods have been provided to address many individual tasks, for example \cite{Ahn2020,Cabrera2020} for subgroup fairness, \cite{amershi2015modeltracker} for identifying problematic items, and \cite{Wexler2020whatif,gomezViCEVisualCounterfactual2020} use counterfactuals to identify issues in model performance. \cite{hypoml} considers designing experiments to assess details in a rigorous manner. Boxer \cite{boxer} shows how a unified system can address a range of detailed classifier assessment tasks by providing a coordination interface
based on 
the selection and comparison of subgroups.

Our approach integrates specific support for trinary tasks (e.g., calibration and operating point selection) with detailed assessment tools. These two complement each other: detailed assessment can be used in the trinary tasks (e.g., identifying examples of confident errors) and the trinary results can drive detailed exploration (e.g., examining the rejected results as interesting examples). In principle, our views and tools for trinary tasks could integrate with many of the approaches for detailed assessment discussed above. In practice, we have integrated our approach with the selection-comparison interface introduced in \cite{boxer}. We discuss this integration in the next section.

\section{Tasks and Approach for Continuous Classifier Assessment}
\label{sec:tasks}
\boxerbasicsfig

\added{We divide the workflow of exploring classifier results into three phases: \emph{calibration} where the scores are checked to confirm that they are meaningful, \emph{operating point selection} where thresholds are chosen to group examples into discrete categories, and \emph{examination} where the results are considered in detail. The phases are initially done in order: it is not worth setting thresholds if the scores are not determined to be meaningful, and thresholding into groups is a useful tool for subsequent analysis. However, they are often iterative and interconnected. For example, results of an analysis might lead to re-examination of calibration and thresholds, or specific examples may be examined in the calibration process.}

\added{Each phase may involve different basic tasks. Practitioners assess and compare classifiers to determine if they are good enough and to select among them. They examine performance for different hyper-parameter values (including thresholds) so select and adjust them. And they consider results to guide tuning and improvement.}
\deleted{Practitioners perform several basic tasks with the results of continuously valued classifiers. Many of these tasks are aggregate. They assess and compare performance to determine whether a classifier is sufficiently good or select which is best. They assess and compare calibration to see if the scores of the classifier have meaning. They examine performance at different operating points to select these thresholding parameters. }
Practitioners have a range of statistical measures and graphs that address these aggregate tasks. However, these aggregate tasks also lead to more focused analysis. Practitioners need to identify different subgroups or exemplars where some property (e.g., performance, calibration) is noteworthy, to check for fairness, to identify and diagnose potential problems, or to learn more about their data.
These tasks often involve comparison between classifiers, between groups, or between classifiers focused on specific groups.

Our approach extends tools used for classifier assessment to include the comparison and group operations required to support tasks.
We enhance standard views
to support comparison, showing multiple classifiers and/or their statistics over selected subgroups. We also extend the views to allow for flexible selection of groups of items of interest, such that these groups can be viewed and compared in multiple ways.
For certain tasks, we find the existing views insufficient and introduce new ones.

We introduced the select and compare approach for classifier examination in the \boxer system \cite{boxer}.
In that prior work, we provided approach for discrete choice classifiers. In this paper, we apply the approach to continuous ones.
The approach allows the results of multiple classifiers on a testing set to be loaded and treated as a single data set that can be explored. The system provides a variety views of this data that can be instantiated and combined on demand.
The system provides mechanisms where the user can select subsets of items. The selections appear across views providing connection, can serve to focus exploration, and can be refined. The ability to specify multiple selections allows for making comparisons between subgroups of interest and for refining the selections by combining them with set arithmetic. \autoref{fig:basicboxes} illustrates the basic \boxer mechanisms.
Standard views include various forms of confusion matrices and stacked bar charts to summarize classifier performance into selectable subgroupings. Any bar or matrix square can be clicked to select their corresponding item set, and these selections can be combined using set algebra. The overlap between selections and each bar or square is shown.

Continuously valued classifiers offer new challenges that we address. First, they offer the new tasks of calibration and operating point selection. In \autoref{sec:tuning}, we discuss how we add views that specifically support these tasks, both enhancing existing views and introducing novel views. We describe how we support exploratory interactions for operating point selection, and workflows where multiple operating points are compared. We also discuss how we can partially mitigate issues around generalization by encouraging users to work with samplings of the dataset. A second type of challenge involves extending the select-and-compare assessment approach to work with continuously valued data. In \autoref{sec:detail} we describe views that use trinary classification to enable selection, comparison, and view composition, even if the goal is not a rejection classifier.
We describe how we enhance the interactive approach to better support exploration of details and situations with skewed data distributions.

\sysname necessarily provides many different views because it must provide the summary visualizations used by practitioners \emph{and} the new, task-specific views we have designed. The novelty in the system is not in the individual views, but rather how they are designed to work together in a coordinated fashion using the selection mechanisms.

\section{Calibration and Tuning}
  \label{sec:tuning}
We provide views and interactions that support \added{the first two phases of the workflow:} assessment of calibration and the selection of operating points. Calibration is the problem of determining that the scores of a classifier correctly correlate with its performance, while operating point selection is the problem of choosing thresholds for making decisions based on the calibrated outputs. Central to our approach is that these are integrated into the overall select-and-compare mechanisms, allowing for detailed exploration as part of these tasks. Rather than viewing these as separate steps, we view calibration, operating point selection, and detailed analysis as integrated parts of the whole analysis process. The innovation in \sysname is not any particular view, but rather how each view is adapted to work with our coordination mechanisms.

\subsection{Calibration}

Calibration involves assessing the correlation between item scores and classifier performance. Reliability or calibration curves are a standard approach for assessing these \cite{predictingGoodProbabilities2005,scikit-learn,degrootComparisonEvaluationForecasters1983}. Our \reliabilitycurveview (\autoref{fig:teaser}A) implements this view for assessing the calibration of a classifier, plotting a line graph of score (x-axis) vs. percent correct (y-axis), a line with slope 1 represents a calibrated classifier. The view also includes a bar chart showing the number of items for each level of score (the items are binned based on score). The bar chart can be used for selections.

While the \reliabilitycurveview
extends the standard view for assessment of calibration to comparison of multiple classifiers,
it does not necessarily enable more detailed understanding. We include a bar chart of score levels that allows for different levels to be selected for analysis. However, for closer examination, the \perfconfview (introduced in \autoref{sec:perf-conf}) provides a more detailed breakdown of performance in a manner that bring performance and number of items together. Similarly, the \rejectview (discussed below) allows for assessing calibration in the context of how it will be used to create rejection trade-offs around a chosen threshold. Well calibrated classifiers provide improved performance for various rejection levels.

\subsection{Operating Point Selection}
  \label{sec:threshold}

The choice of operating point (i.e., the decision threshold and (optionally) rejection threshold) is often a manual choice as the trade-offs between error types, and interpretation of the errors rates over the testing data, are situation dependent. Our approach is to empower users with views that show performance tradeoffs over the testing data, coupled with methods that allow them to explore the impact of different choices coupled with detailed examination of the results to appreciate how representative they are.
\replaced{This can complement approaches where users model their preferences with cost functions. It can also help in building reliable models of how costs over testing data generalize.}
{This can complement approaches where users model their preferences with cost functions and to build reliable models of how costs over testing data generalize.}

Central to our approach is allowing users to explore the effects of threshold settings interactively.
The \probcontrolpanel (\autoref{fig:teaser}G,H)allows the user to adjust the thresholds using sliders.
Views across the system update in real time as adjustments are made, allowing the user to experiment with changes and understand their ramifications.
Our system provides many different views to summarize performance, including tables of metrics, confusion matrices, and distributional bar charts (see \cite{boxer} for a list), providing flexibility in how these changes are assessed. Views for assessing trinary classifications (\autoref{sec:detail}) are designed to help quickly assess rejection performance.

We include standard trade-off curve views for visualization of performance trade-offs over the sample data.
The \perfcurveview shows Receiver Operating Characteristic (ROC) and Precision/Recall curves for multiple classifiers and selections. Similarly, the \rejectview shows the Accuracy-Rejection Curve (ARC) \cite{nadeemAccuracyRejectionCurvesARCs2009,hanczarPerformanceVisualizationSpaces2019} which plots performance against rejection rate (percentage of items rejected), providing access to the threshold levels for different rates of rejection.
The \rejectview extends standard ARCs by permitting the use of a variety of metrics beyond accuracy to help the user understand performance trade-offs (e.g., recall rates in scenarios with unequal costs, MCC to address class imbalance, or the weighted metrics of \autoref{sec:weighted} to emphasize important groups).
The \rejectview can display multiple curves to allow for comparison between classifiers, different decision threshold levels, or different subgroups.

The \perfcurveview and \rejectview allow thresholds to be set directly by choosing points on the trade-off curves. Once an operating point is selected, it can be ``set'' as a new classifier, allowing the full machinery of comparison to be applied to compare between different operating points of the same classifier.

\bandassessfig

The \perfcurveview and \rejectview display trade-offs for the decision and rejection thresholds independently, providing limited ability to visualize the trade offs in the two dimensional space where the thresholds interact.
We introduce two views specifically designed to help with assessing the range of dual thresholds. The \bandwidthassessview (\autoref{fig:bandassess} and \autoref{fig:teaser}D) shows a graph of a performance metric (e.g., accuracy) for the range of threshold values with indication of the effects of different bandwidths. The effects of the bandwidth are shown as glyphs in the accuracy graph: the upper mark indicated the accuracy if the rejected items are considered as correct predictions, the lower mark if the rejected items are considered incorrect. These ``error-bar-like'' glyphs allow the viewer to see the range of effect possible with different levels of bandwidth, effectively converting the amount of rejection to a range of performance values.
Details of the view are discussed in the caption of \autoref{fig:bandassess}.


The \uncertheatview (\autoref{fig:teaser}C) shows performance for various values of upper and lower thresholds simultaneously. Thresholds are mapped to position, a performance metric (e.g., accuracy or recall) is mapped to color, and the percentage of non-rejected items is mapped to radius. Settings that reject many items are less salient --- even if they have favorable colors. Hovering over a circle exposes the numerical values for all metrics. The view allows for quick identification of favorable ranges in terms of performance, while considering rejection rate. The color map is designed to be legible event with small dot sizes \cite{szafirModelingColorDifference2018}, using a colorful, iso-luminant ramp for the relevant part (values above chance), and a color to gray ramp for other part.
\added{For example, \autoref{fig:teaser}C shows how the \uncertheatview} provides a quick overview of the range of upper and lower thresholds. We can see that high accuracy is achievable (yellow dots), but these high accuracy settings lead to higher rejection rates (smaller dots, represented fewer non-rejected samples). The large orange along the diagonal suggest tradeoff situations with relatively high accuracy but low rejection rates.

\subsection{Addressing Generalization}
\label{sec:sampling}
Calibration and operating point selection for learned classifiers are \emph{empirical} processes. They require examining the performance of the classifiers over sets of testing data. This empirical nature poses problems for analysis. First, the testing data is part of the experiment: analysis may need to connect to the data to help interpret how representative the testing set is of the population the classifier will ultimately be deployed on. Second, because the testing set is only a sample of the population, analysis must try to make decisions that are likely to generalize to the broader population.

Overfitting is a common problem in machine learning where a model is specialized to the details of its training set. A similar problem can occur in  assessment: the user can inadvertently draw conclusions or identify patterns that are specific to the testing set. The problem is literally overfitting when the user sets hyper-parameters (e.g., thresholds) based on their exploration, but can occur more subtly as the user draws conclusions from the given testing set.

Practitioners must be cautious when performing detailed assessment to make observations and adjustments that generalize beyond the test data.
\sysname is designed to support standard practices for to splitting data.
For example, \sysname allows users to explore on the training set used to build the classifiers, and then only examine the actual testing set to assess their findings.
Using the training set may be problematic because the resubstitution errors (errors applying a classifier to its training data) may have different properties than the test set.

Another standard practice is to use sampling to create subsets of the data, allowing for synthetic experiments. \sysname supports this practice with facilities for creating random groupings of the data.
This allows a user to draw observations from one partition and then assess their generality on the other. The non-selected partition can be completely hidden, or can be accessed as a separate selection, allowing for exploration that excludes the set, or comparison between the test and validation sets.
Our partition tool can either perform random sampling or stratified sampling.

\sysname can also create variants of the data set by \emph{bootstrap sampling}.
Examining such variants allows for a user to check that properties apply to the likely distribution the data is drawn from. For example, thresholds can be determined for different variants to estimate a confidence interval.
The sampling tools cannot prevent misuse of detailed observations, but they facilitate the use of common responsible practices.

\section{Detail Assessment with Selection and Comparison}
\label{sec:detail}

We extend the select-and-compare approach of \boxer \cite{boxer} to address continuous classifier values. Some of these changes are straightforward. For example, summary views, such as tables and parallel coordinates displays of metrics, are enhanced with continuous summary metrics including AUC and MSE.
We introduce new views specifically designed for comparative display and selection of trinary data (\autoref{sec:compare-views}). We also introduce features that better enable the approach to handle skewed data distributions (\autoref{sec:weighting}) and examination of item details (\autoref{sec:focus-item}). \added{Most views in the system provide mechanisms for specifying and displaying dual selections, as discussed in \cite{boxer}.}

\subsection{Selection Mechanisms}
\label{sec:compare-views}
Central to the select-and-compare approach are views that enable selecting relevant sets of items, and displaying how selected sets are distributed. In addition to the views provided for discrete classifiers \cite{boxer}, the \reliabilitycurveview and \bandwidthassessview provide bar charts of distributions for selection and display. Additionally we introduce two new views.

\label{sec:perf-conf}
The \perfconfview (\autoref{fig:teaser}B) shows a histogram (bar chart) of the number of items for each binned level of score. Stacked bars show different classifications at each level. This basic design uses a position encoding for score level and color for correctness as introduced by \cite{amershi2015modeltracker}. We use a vertical design that better uses space when juxtaposing classifiers for comparison. We bin the items (rather than providing individual marks) for scalability, relying on selection mechanisms to identify individuals. The stacked bar charts divide the actual class into areas; the prediction is encoded by position relative to the threshold. Our current design uses colors for correct and incorrect: negative/positive must be inferred by position relative to the threshold. Experiments with separate colors led to an even more busy display.

\trinaryviewfig
The \trinaryview (\autoref{fig:trinary} and \autoref{fig:teaser}E) provides a summary of the classification decisions for given threshold values, allowing for assessment of a threshold setting. The standard confusion matrix does not work as there is a 5th category (rejected items). Instead, the design uses a stacked bar design to allow for comparison between classifiers. To enable different comparisons against common baselines, the ordering of the stacked elements can be changed. Correctness mode groups right and wrong answers, while score mode groups positive and negative predictions.

\subsection{Handling Distributions}
\label{sec:weighting}
\label{sec:weighted}
In many data sets, small groups are relevant. For example, there may be a skewed distribution of data (i.e., positives are rare), or we may be interested in understanding the performance over a small sub-population. The \sysname interface includes features to facilitate working with small groups and skewed distributions. Small bars in bar charts are given circular glyphs that indicate small, but non-zero, quantities with a visible and selectable amount of screen space. Several charts can be limited to show distributions over selections, and performance displays allow selections to compared.

\sysname also supports weighting instances based on groups. Weights are used in computing performance metrics, such as accuracy, precision, recall, F1, and MCC. Any selection can have a weight applied. For example, selections of classes allow for weighting to address class imbalance, or selecting a subgroup based on a data attribute allows for increasing the importance of a sensitive group. Weights for different selections can be combined, for example to provide weights for both classes as well as other data groupings.

\subsection{Seeing items across the interface}
\label{sec:focus-item}
Our system design focuses on summary views that scale to handle large testing sets. However, identifying specific data items is an important part of the approach.
\sysname extends our prior approach with features for identifying and examining specific items.
\sysname has a list view that shows data items in a tabular format, allowing for detailed examination of selected items. We have also added a \scatterview (\autoref{fig:scatter}) that shows how the items distribute over two user selected continuous variables, such as classifier scores or a feature variables. The \scatterview scales to large item collections by automatically switching to a color encoding of binned density \cite{heimerlVisualDesignsBinned2020}. However, it also shows small sets of selected items overlayed on this density field to provide context, and allows for zooming into regions to see individual points.
\scatterfig

We have also introduce the notion of a \emph{focus item}: a single item that is specially selected (independent of the two set selections) that is shown throughout the interface. A specific \focusview shows the details of this item, for example showing the image (if the data items are images) and any features that are available. The focus item is shown throughout the interface as a yellow dot that appears in most views indicating where the focus item fits in the various distributions displayed. The focus item can be selected using the list view, the \scatterview. However, the \focusview includes controls for  selecting items either in order or randomly from either the selections of the entire data set. This allows for quickly scanning over selections, or finding random examples to examine in sequence.

\section{Use Cases}

This section provides example scenarios where the views and interactions of our approach are combined to address the kinds of tasks that we see practitioners request support for. The data sets in \autoref{sec:intro-example}, \autoref{sec:uc1} and \autoref{sec:uc2} are benchmark sets from \cite{uci} with classifiers built using Scikit-learn \cite{scikit-learn}. \autoref{uc:cifar} provides an example with a standard computer vision benchmark and classifiers created using deep neural networks. \autoref{sec:uc3} is an example of using classifier analysis as a tool to help domain collaborators explore their data.

All examples were completed using the \sysname prototype system that implements our approach. \sysname is a web application implemented in TypeScript using the Vue.js application framework and D3.js for drawing visualizations. It loads all data, including classification results, item features, and meta-data, into memory at startup and performs all computations within the browser. The classification results are pre-computed and provided as input data. 

\subsection{Model Selection} 
	\label{sec:uc1}

\label{sec:income-modelsel}
\ucincome

This use case demonstrates how calibration benefits from detailed assessment.
It shows how integrating standard aggregate views for comparison and detail examination enables selection based on calibration.
The scenario uses the income classification benchmark dataset from \cite{uci} that has been downsampled.
Classifiers determine whether an individual's income is above a certain level. Three models were constructed with different methods:
logistic regression (LR) and Na\"ive Bayes (NB). The models provide very similar correctness summary scores (F1 is .783, identical to 3 decimal places) over a test set of 1500 items. The example is shown in \autoref{fig:income-modelsel}.

We prefer a calibrated classifier \added{because it allows us to use the scores to balance error types or assess confidence}. The AUC score suggests LR is slightly better, which is confirmed by the \reliabilitycurveview. However this view also shows that there are few items with middle scores, so the curve may be unreliable. We choose to examine the differences in detail to confirm and explore the differences. The \perfconfview (\autoref{fig:income-modelsel}C) shows a clear picture: NB provides a sharp distribution \cite{gneitingProbabilisticForecastsCalibration2007} with many extreme values. We see that many of the errors have extreme values. In contrast, LR has many scores in the middle, and many of the errors are there as well.

To explore the difference, we select the items that NB has errors and extreme scores by unioning together the error in the highest and lowest bars of the \perfconfview (\autoref{fig:income-modelsel}C).
While LR also gets many of these wrong, its errors tend to be distributed in the middle scores. Moreover, if we want to compare performance by considering rejections, \rejectview shows the F1 score of LR raises above .9 when the rejection rate is fixed around 20\%, while the  F1 for NB is effectively unchanged.
We obtain thresholds for LR between .4 and .6 using the \uncertheatview (\autoref{fig:income-modelsel}E). We see this eliminates the majority of LR errors.


\subsection{Hyper parameter Tuning: Threshold} 
  \label{sec:uc2}
\ucwinefig
\label{sec:wine-threshold}

This use case illustrates how detailed comparisons can improve threshold selection.
We see how the ability to make comparisons of details can improve understanding threshold choices.
The scenario uses wine quality benchmark from \cite{uci} which requires classifying the quality of a wine from its properties. Because we are trying to determine a hyper-parameter, we perform this analysis using the training data. We build a random forest classifier, and find that it gives good performance over a range of thresholds. We want to understand if changes in the threshold affect the outcomes.

Using the \bandwidthassessview, we see that the accuracy is constant for thresholds around .5. We use the \probcontrolpanel to adjust the threshold. For several values in the range, we create new classifiers that use the same model with different thresholds (.5, .55, .6, .65) and compare these models. We see that accuracy, F1 and MCC scores are very similar across the four. However,  selecting the errors show different details: for the lower thresholds, there are more false negatives, while for the higher thresholds more false positives (\autoref{fig:uc2-wine-threshold}). We see that this classifier is quite sensitive to small changes in the threshold, even though those changes do not affect the most common metrics.

\subsection{Model Selection and Detail Examination}
\label{uc:cifar}
\label{sec:cifar}

This use case shows how our approach can help with model selection and threshold tuning.
We use the integration of views to understand differences in models and to integrate tuning into selection, we also use the ability to examine detailed groups to better understand performance.
A classifier was built for the CIFAR 100 \cite{krizhevsky2009learning} computer vision benchmark using Tensorflow \cite{tensorflow2015-whitepaper}. The data set has 100 classes, and the trained classifier produces a distribution over these classes as its decision. Our goal is to create a binary classifier for a ``meta-class'' which combines 5 of the main classes. In particular, we want to classify flowers, which can be any one of 5 of the original classes. Because the test set contains all 100 classes, it is quite imbalanced: flowers are only 5\% of the total instances. The results are shown in \autoref{fig:cifar}.

\cifarfig

Classifiers were created using three different strategies that combine the base classes: \emph{sum}, \emph{average}, and \emph{largest}. Because of the class imbalance, we use Mathews Correlation (MCC) as the metric. Each combination strategy produces different ranges of scores. We can use the \reliabilitycurveview (\autoref{fig:cifar}A) to see the differences, and estimate appropriate thresholds for each.
For each model, we use the \bandwidthassessview (\autoref{fig:cifar}B) to choose a threshold that provides high MCC yet provides a range of available rejection rates. The \rejectview (\autoref{fig:cifar}C) shows that each model gets a performance gain from a 10\% rejection rate. After tuning each model appropriately all have similar performance.

Using the \trinaryview (\autoref{fig:cifar}E) we can make selections to look for differences in the similar performance. We note that while each model rejects the same number of items, they reject different items. While the number of errors is small, they are different between models. One model has more false positives, while the other has more false negatives.

We select the false positives of the \emph{largest} model. The \focusview (\autoref{fig:cifar}F) allows us to step through these errors to look for patterns. We notice that many of the errors are flowers with insects on them. Because these images are labeled as insect, they are scored as misclassified.

\subsection{Data Examination}
  \label{sec:uc3}

This use case considers using \sysname to learn more about our data.
We use the flexibility of examining scores in aggregate and over specific groups to check theories against data, and use sampling to check results.
We use a corpus of 59,989 documents from a historical literary collection: Text Creation Partnership (TCP) transcriptions of the Early English Books Online (EEBO), i.e., transcriptions of books published between 1475-1700.
The data counts the 500 most common English words in each document.
While all documents have been classified by experts, we construct classifiers using the data to support theories that different types of documents use words in different ways~\cite{Witmore2010,Gleicher2013}.

We construct classifiers that determine whether a document is written after 1642 based on the 500 most common words in the corpus.
While ground truth is known, effective classifiers can help understand how word usage changed at this critical date that marks the beginning of the English Civil War.
The collection is skewed (only 25\% of the documents were written before 1642).
Therefore, we prefer MCC as a correctness metric.

For the experiment, we took a random sample of 2500 documents, and held out 30\% using stratified sampling.
We constructed random forest (RF) and logistic regression classifiers (LR), with and without applying corrections for class skew.
The random forest classifiers achieve perfect
resubstitution 
for a wide range of thresholds,
but worse performance with the default (.5) threshold on the test set.
However, after choosing the threshold that achieves the highest MCC for each classifier, the RF classifiers have the best performance.
Using the \bandwidthassessview we can choose the threshold that optimizes MCC for each classifier.
After this optimization, the RF classifiers have superior performance across metrics.
However, this optimized performance may be specific to the test set. To provide a check against overfitting, we use CBoxer's bootstrap sampling feature to create 10 new testing sets from the original and see that the results do not change. While the optimal values of threshold and metrics may change slightly, in all cases, RF beats LR.
\ucdatesperf

A theory in literary linguistics is that the civil war marked a radical and rapid change in written language. To help assess this theory, we can examine if there is a clean division between before and after given the data. We can select the time periods before and after the critical date and see that performance is poor. Scores are generally close to the threshold (\autoref{fig:lit-dates}) are neither clearly classified as either before or after. This is more consistent with a gradual transition and that the change in language began earlier.


Another question is whether the length of documents has any connection to the certainty of a prediction. Using a \histogramview, we select the shortest and longest documents, and view the distribution over performance confidence view (\autoref{fig:lit-lengths}). We can see that long documents are disproportionately given confident scores, while the shortest documents (less than 150 words) score in the center (less confident predictions).
\ucdatesfiglengths

\section{Discussion}
\label{sec:discussion}

This paper has provided an interactive approach for addressing key tasks examining the results of continuously valued binary classifiers. We provide views for calibration and threshold setting that enhance standard summaries by supporting comparison and selection. We extend the select-and-compare approach for classifier result examination by introducing trinary classification (positive/unsure/negative) even in cases where the ultimate classifier will be binary. We have integrated these ideas in an open source prototype system and used it to show the promise of our approach. While our initial implementation in \sysname shows the promise of the approach, it also makes us aware of a number of limitations in the present work.

\noindent\textbf{Scope: extending to other problem types.} Our work focuses on binary classification. Extending to a multi-class setting is challenging as the scores become vectors. These vectors can be converted to decisions in many ways (e.g., winner-take all, softmax, majority required) so the notion of thresholds may not apply. Correctness can be considered in many ways (e.g., single correct choice, answer in top-N, per-class binary). Designs for considering multi-class probabilities, e.g., \cite{ren2017squares,alsallakh2014visual}, could be generalized and adapted to fit within our comparative framework. The notion of rejection becomes more complex in the multi-class setting, as the rejection criteria may depend on the distribution of scores.
Our approach focuses on situations where the classification score is used to determine the rejection criteria. Extending to situations where there is an independent rejection score would require our views to be updated to show the distributions of both scores.

\noindent\textbf{Scalability: extending to larger problems.} Scalability must consider all three axes of hardness \cite{Gleicher2018}: large numbers of test items, large numbers of classifiers to compare, and complex patterns. Our visual designs scale well to large test sets, as they provide aggregate views and features that help spot small groups within large ones; however, our current in-browser implementation is not sufficiently performant to provide interactive manipulation of thresholds when large item sets are used. Our designs scale less well to comparisons between many classifiers.

\noindent\textbf{Addressing Generalization:} The goal of most predictive modeling is to create models that generalize beyond the current data. Focusing on the testing set can lead users to over-generalize their observations, for example to overfit their ideas to the training set. Our sampling features are a first step in improving this situation, we need to better consider how to help users avoid analytic pitfalls. Beyond richer sampling strategies, we also envision guidance that can warn users against mistakes. We also plan to introduce statistical analysis, as in \cite{hypoml}, so that users can netter assess generalization.

\noindent\textbf{Usability:} A significant drawback of our approach is its complexity: a user must determine how to best combine our views and interaction mechanisms to achieve their goals. Goals are rarely well-defined and tend to evolve during analysis, which exacerbates this challenge. We are seeking to address the usability challenge through the use of pre-configured layouts to answer specific questions (as in~\cite{Szafir2016textDNA}) and workflow-based guidance to suggest potential next steps to a user~\cite{Ceneda2017}.
\added{While there is a typical high-level workflow (calibrate, select operating points, assess), the details of these workflows depend on task, situation, and user preference. While we provide exemplar workflows in the user guide, these are less helpful in guiding users to exploit the flexibility in the system to address specific scenarios.}

\noindent\textbf{Black-Box Analysis:} Our focus on black-box analysis makes our approach general across model types and useful to users who are not concerned with how models work. However, the ability to connect classifier results to model internals can help developers tune and debug their models.

\noindent\textbf{Integrated pipelines:} \sysname is presently a standalone tool for assessing the results of constructed classifiers.
Integrating its ideas into a comprehensive approach to modeling, allowing for iteration and feedback loops, would better support users in model development and debugging tasks. A particular challenge will be integrating the highly interactive, multi-view coordination required to support the select-and-compare approach with the notebook-style workflows favored by many practitioners\cite{keryMageFluidMoves2020}. At present, we provide scripts that allow classifier data to be exported from notebooks to \sysname; future tighter integration should allow for automatic updating and bi-directional communication.

\added{\noindent\textbf{Fixing Problems:} \sysname works in an ``open loop'' fashion: it can help a user identify problems, and provide information such as exemplars that might be useful in addressing them. However, it does not directly provide support for improving the classifier. For example, \sysname can help the user observe that a classifier is uncalibrated, and its scores should not be used as they have little meaning, but it does not help fix this problem.}

\noindent\textbf{Evaluation:} Our system has had limited use beyond the development team. While this has given us to opportunity to validate our designs in example use cases, we have not been able confirm the systems utility by actual practitioners.
Informal feedback from work with potential users has improved our designs.

\subsection{Conclusion}
Despite its limitations, our approach and implementation can already address many problems faced by machine learning practitioners. We are continuing to work with them to better understand their needs and adapt tools to meet them.

Our work shows that we can support examination tasks with continuously-valued binary classifiers using a visual-interactive approach.
\added{We combine support for confirming that the classifier output scores are meaningful into support for using the scores.}
To do this, we have extended standard summary views as well as introduced new, task-specific views. These extensions enable the views to work within a select-and-compare framework, affording the use of coordination to enable detailed exploration and comparison. The extensions are unified through their use of trinary classification, which proves useful even if the ultimate classifier will be binary.

\section*{Acknowledgments}
This research was supported in part by NSF awards 1841349 and 2007436.

\ifieee
\else
\bibliographystyle{elsarticle-num}
\bibliography{boxer_no_url}

\begin{thebibliography}{10}
\expandafter\ifx\csname url\endcsname\relax
  \def\url#1{\texttt{#1}}\fi
\expandafter\ifx\csname urlprefix\endcsname\relax\def\urlprefix{URL }\fi
\expandafter\ifx\csname href\endcsname\relax
  \def\href#1#2{#2} \def\path#1{#1}\fi

\bibitem{chowOptimumRecognitionError1970}
C.~K. Chow, On optimum recognition error and reject tradeoff, IEEE Trans. Inf.
  Theory (1970).

\bibitem{landgrebeInteractionClassificationReject2006}
T.~C.~W. Landgrebe, D.~M.~J. Tax, P.~Pacl{\'i}k, R.~P.~W. Duin, The interaction
  between classification and reject performance for distance-based
  reject-option classifiers, Pattern Recognition Letters 27~(8) (2006)
  908--917.

\bibitem{condessaPerformanceMeasuresClassification2017}
F.~Condessa, J.~{Bioucas-Dias}, J.~Kova{\v c}evi{\'c}, Performance measures for
  classification systems with rejection, Pattern Recognition 63 (2017)
  437--450.

\bibitem{uci}
D.~Dua, C.~Graff, {UCI} machine learning repository (2017).

\bibitem{Cabrera2020}
A.~A. Cabrera, W.~Epperson, F.~Hohman, M.~Kahng, J.~Morgenstern, D.~H. Chau,
  Fairvis: Visual analytics for discovering intersectional bias in machine
  learning, IEEE Conference on Visual Analytics Science and Technology (VAST)
  26~(1) (2020).

\bibitem{Ahn2020}
Y.~{Ahn}, Y.~{Lin}, Fairsight: Visual analytics for fairness in decision
  making, IEEE TVCG 26~(1) (2020) 1086--1095.

\bibitem{provostCaseAccuracyEstimation1998}
F.~J. Provost, T.~Fawcett, R.~Kohavi, The {{Case}} against {{Accuracy
  Estimation}} for {{Comparing Induction Algorithms}}, in: Proceedings of the
  {{Fifteenth International Conference}} on {{Machine Learning}}, {{ICML}} '98,
  {Morgan Kaufmann Publishers Inc.}, {San Francisco, CA, USA}, 1998, pp.
  445--453.

\bibitem{Powers2011}
D.~Powers, {Evaluation: From Precision, Recall and F-Measure to ROC,
  Informedness, Markedness and Correllation}, Journal of Machine Learning
  Technologies 2~(1) (2011) 37--63.

\bibitem{parkerAnalysisPerformanceMeasures2011}
C.~Parker, An {{Analysis}} of {{Performance Measures}} for {{Binary
  Classifiers}}, 2011 IEEE 11th International Conference on Data Mining (2011).

\bibitem{chiccoAdvantagesMatthewsCorrelation2020}
D.~Chicco, G.~Jurman, The advantages of the {{Matthews}} correlation
  coefficient ({{MCC}}) over {{F1}} score and accuracy in binary classification
  evaluation, BMC Genomics (2020).

\bibitem{lingAUCBetterMeasure2003}
C.~Ling, J.~Huang, H.~Zhang, {{AUC}}: {{A Better Measure}} than {{Accuracy}} in
  {{Comparing Learning Algorithms}}, in: Canadian {{Conference}} on {{AI}},
  2003.

\bibitem{brierVerificationForecastsExpressed1950}
G.~W. Brier, Verification of forecasts expressed in terms of probability,
  Monthly Weather Review 78~(1) (1950) 1--3.

\bibitem{gneitingProbabilisticForecastsCalibration2007}
T.~Gneiting, F.~Balabdaoui, A.~E. Raftery, Probabilistic forecasts, calibration
  and sharpness, Journal of the Royal Statistical Society: Series B
  (Statistical Methodology) 69~(2) (2007) 243--268.

\bibitem{schwarz2019guess}
J.~Schwarz, D.~Heider, Guess: projecting machine learning scores to
  well-calibrated probability estimates for clinical decision-making,
  Bioinformatics 35~(14) (2019) 2458--2465.

\bibitem{guoCalibrationModernNeural2017}
C.~Guo, G.~Pleiss, Y.~Sun, K.~Q. Weinberger, On {{Calibration}} of {{Modern
  Neural Networks}}, in: International {{Conference}} on {{Machine Learning}},
  {PMLR}, 2017, pp. 1321--1330.

\bibitem{Kuleshov2015_CalibratedStructedPrediction}
V.~Kuleshov, P.~S. Liang, Calibrated structured prediction, in: C.~Cortes,
  N.~Lawrence, D.~Lee, M.~Sugiyama, R.~Garnett (Eds.), Advances in Neural
  Information Processing Systems, Vol.~28, {Curran Associates, Inc.}, 2015, pp.
  3474--3482.

\bibitem{zhongAccurateProbabilityCalibration2013}
L.~W. Zhong, J.~T. Kwok, Accurate probability calibration for multiple
  classifiers, in: Proceedings of the {{Twenty}}-{{Third}} International Joint
  Conference on {{Artificial Intelligence}}, {{IJCAI}} '13, {AAAI Press},
  {Beijing, China}, 2013, pp. 1939--1945.

\bibitem{plattProbabilisticOutputsSupport1999}
J.~C. Platt, Probabilistic {{Outputs}} for {{Support Vector Machines}} and
  {{Comparisons}} to {{Regularized Likelihood Methods}}, in: Advances in
  {{Large Margin Classifiers}}, {MIT Press}, 1999, pp. 61--74.

\bibitem{lingBayesianClassifiersAccurate2002}
C.~X. Ling, H.~Zhang, Toward {{Bayesian Classifiers}} with {{Accurate
  Probabilities}}, in: Proceedings of the 6th {{Pacific}}-{{Asia Conference}}
  on {{Advances}} in {{Knowledge Discovery}} and {{Data Mining}}, {{PAKDD}}
  '02, {Springer-Verlag}, {Berlin, Heidelberg}, 2002, pp. 123--134.

\bibitem{jiangTrustNotTrust2018}
H.~Jiang, B.~Kim, M.~Guan, M.~Gupta, To {{Trust Or Not To Trust A Classifier}},
  Advances in Neural Information Processing Systems 31 (2018) 5541--5552.

\bibitem{cortesLearningRejection2016}
C.~Cortes, G.~DeSalvo, M.~Mohri, Learning with {{Rejection}}, in: R.~Ortner,
  H.~U. Simon, S.~Zilles (Eds.), Algorithmic {{Learning Theory}}, Lecture
  {{Notes}} in {{Computer Science}}, {Springer International Publishing},
  {Cham}, 2016, pp. 67--82.

\bibitem{predictingGoodProbabilities2005}
A.~{Niculescu-Mizil}, R.~Caruana, Predicting good probabilities with supervised
  learning, in: Proceedings of the 22nd International Conference on {{Machine}}
  Learning, {{ICML}} '05, {Association for Computing Machinery}, {New York, NY,
  USA}, 2005, pp. 625--632.

\bibitem{scikit-learn}
F.~Pedregosa, G.~Varoquaux, A.~Gramfort, V.~Michel, B.~Thirion, O.~Grisel,
  M.~Blondel, P.~Prettenhofer, R.~Weiss, V.~Dubourg, J.~Vanderplas, A.~Passos,
  D.~Cournapeau, M.~Brucher, M.~Perrot, E.~Duchesnay, Scikit-learn: Machine
  learning in {P}ython, Journal of Machine Learning Research 12 (2011)
  2825--2830.

\bibitem{leathart2020temporal}
T.~Leathart, M.~Polaczuk, Temporal probability calibration, arXiv preprint
  arXiv:2002.02644 (2020).

\bibitem{hanczarPerformanceVisualizationSpaces2019}
B.~Hanczar, Performance visualization spaces for classification with rejection
  option, Pattern Recognition 96 (2019) 106984.

\bibitem{nadeemAccuracyRejectionCurvesARCs2009}
M.~S.~A. Nadeem, J.-D. Zucker, B.~Hanczar, Accuracy-{{Rejection Curves}}
  ({{ARCs}}) for {{Comparing Classification Methods}} with a {{Reject Option}},
  in: Machine {{Learning}} in {{Systems Biology}}, {PMLR}, 2009, pp. 65--81.

\bibitem{Gleicher2016}
M.~Gleicher, {A Framework for Considering Comprehensibility in Modeling}, Big
  Data 4~(2) (2016) 75--88.

\bibitem{Lipton2016}
Z.~C. Lipton, {The Mythos of Model Interpretability}, arXiv preprint
  arXiv:1606.03490 (Jun 2016).
\newblock \href {http://arxiv.org/abs/1606.03490} {\path{arXiv:1606.03490}}.

\bibitem{amershi2015modeltracker}
S.~Amershi, M.~Chickering, S.~M. Drucker, B.~Lee, P.~Simard, J.~Suh,
  Modeltracker: Redesigning performance analysis tools for machine learning,
  in: Proceedings of the 33rd Annual ACM Conference on Human Factors in
  Computing Systems, 2015, pp. 337--346.

\bibitem{Krause2016}
J.~Krause, A.~Perer, K.~Ng, Interacting with {{Predictions}}: {{Visual
  Inspection}} of {{Black}}-box {{Machine Learning Models}}, in: Proceedings of
  the 2016 {{CHI Conference}} on {{Human Factors}} in {{Computing Systems}} -
  {{CHI}} '16, {ACM Press}, {New York, New York, USA}, 2016, pp. 5686--5697.

\bibitem{ren2017squares}
D.~Ren, S.~Amershi, B.~Lee, J.~Suh, J.~D. Williams, Squares: Supporting
  interactive performance analysis for multiclass classifiers, IEEE TVCG 23~(1)
  (2017) 61--70.

\bibitem{alsallakh2014visual}
B.~Alsallakh, A.~Hanbury, H.~Hauser, S.~Miksch, A.~Rauber, Visual methods for
  analyzing probabilistic classification data, IEEE TVCG 20~(12) (2014)
  1703--1712.

\bibitem{heyen2020clavis}
F.~Heyen, T.~Munz, M.~Neumann, D.~Ortega, N.~T. Vu, D.~Weiskopf, M.~Sedlmair,
  Clavis: An interactive visual comparison system for classifiers, in:
  Proceedings of the International Conference on Advanced Visual Interfaces,
  2020, pp. 1--9.

\bibitem{zhang2018manifold}
J.~Zhang, Y.~Wang, P.~Molino, L.~Li, D.~S. Ebert, Manifold: A model-agnostic
  framework for interpretation and diagnosis of machine learning models, IEEE
  TVCG 25~(1) (2018) 364--373.

\bibitem{Wexler2020whatif}
J.~Wexler, M.~Pushkarna, T.~Bolukbasi, M.~Wattenberg, F.~Vi{\'e}gas, J.~Wilson,
  The what-if tool: Interactive probing of machine learning models, IEEE TVCG
  26~(1) (2020) 56--65.

\bibitem{boxer}
M.~Gleicher, A.~Barve, X.~Yu, F.~Heimerl, Boxer: {{Interactive Comparison}} of
  {{Classifier Results}}, Computer Graphics Forum 39~(3) (2020) 181--193.

\bibitem{das2020questo}
S.~Das, S.~Xu, M.~Gleicher, R.~Chang, A.~Endert, Questo: Interactive
  construction of objective functions for classification tasks, Computer
  Graphics Forum 39~(3) (2020).

\bibitem{Kapoor2010}
A.~Kapoor, B.~Lee, D.~Tan, E.~Horvitz, {Interactive optimization for steering
  machine classification}, in: Proceedings of the 28th international conference
  on Human factors in computing systems - CHI '10, ACM Press, New York, New
  York, USA, 2010, p. 1343.

\bibitem{gomezViCEVisualCounterfactual2020}
O.~Gomez, S.~Holter, J.~Yuan, E.~Bertini, {{ViCE}}: Visual counterfactual
  explanations for machine learning models, in: Proceedings of the 25th
  {{International Conference}} on {{Intelligent User Interfaces}}, {{IUI}} '20,
  {Association for Computing Machinery}, {Cagliari, Italy}, 2020, pp. 531--535.

\bibitem{hypoml}
Q.~{Wang}, W.~{Alexander}, J.~{Pegg}, H.~{Qu}, M.~{Chen}, Hypoml: Visual
  analysis for hypothesis-based evaluation of machine learning models, IEEE
  TVCG 27~(1) (2021).

\bibitem{degrootComparisonEvaluationForecasters1983}
M.~H. DeGroot, S.~E. Fienberg, The {{Comparison}} and {{Evaluation}} of
  {{Forecasters}}, Journal of the Royal Statistical Society. Series D (The
  Statistician) 32~(1/2) (1983) 12--22.

\bibitem{szafirModelingColorDifference2018}
D.~A. Szafir, Modeling {{Color Difference}} for {{Visualization Design}}, IEEE
  TVCG 24~(1) (2018) 392--401.

\bibitem{heimerlVisualDesignsBinned2020}
F.~Heimerl, C.-C. Chang, A.~Sarikaya, M.~Gleicher, Visual {{Designs}} for
  {{Binned Aggregation}} of {{Multi}}-{{Class Scatterplots}}, arXiv:1810.02445
  [cs] (Jan. 2020).
\newblock \href {http://arxiv.org/abs/1810.02445} {\path{arXiv:1810.02445}}.

\bibitem{krizhevsky2009learning}
A.~Krizhevsky, G.~Hinton, Learning multiple layers of features from tiny
  images, Technical {{Report}}, {University of Toronto} (2009).

\bibitem{tensorflow2015-whitepaper}
M.~Abadi, A.~Agarwal, P.~Barham, E.~Brevdo, J.~Dean, {et. al.}, {TensorFlow}:
  Large-scale machine learning on heterogeneous systems, White paper, {Google},
  software available from tensorflow.org (2015).

\bibitem{Witmore2010}
M.~Witmore, J.~Hope, {The Hundredth Psalm to the Tune of “Green Sleeves”:
  Digital Approaches to Shakespeare's Language of Genre}, Shakespeare Quarterly
  61~(3) (2010) 357--390.

\bibitem{Gleicher2013}
M.~{Gleicher}, Explainers: Expert explorations with crafted projections, IEEE
  TVCG 19~(12) (2013) 2042--2051.

\bibitem{Gleicher2018}
M.~Gleicher, {Considerations for Visualizing Comparison}, IEEE TVCG 24~(1)
  (2018) 413--423.

\bibitem{Szafir2016textDNA}
D.~A. Szafir, D.~Stuffer, Y.~Sohail, M.~Gleicher, {TextDNA: Visualizing Word
  Usage with Configurable Colorfields}, Computer Graphics Forum 35~(3) (2016)
  421--430.

\bibitem{Ceneda2017}
D.~Ceneda, T.~Gschwandtner, T.~May, S.~Miksch, H.-J. Schulz, M.~Streit,
  C.~Tominski, {Characterizing Guidance in Visual Analytics}, IEEE TVCG 23~(1)
  (2017) 111--120.

\bibitem{keryMageFluidMoves2020}
M.~B. Kery, D.~Ren, F.~Hohman, D.~Moritz, K.~Wongsuphasawat, K.~Patel, Mage:
  {{Fluid Moves Between Code}} and {{Graphical Work}} in {{Computational
  Notebooks}}, in: Proceedings of the 33rd {{Annual ACM Symposium}} on {{User
  Interface Software}} and {{Technology}}, {{UIST}} '20, {Association for
  Computing Machinery}, {New York, NY, USA}, 2020, pp. 140--151.

\end{thebibliography}
\fi

\newpage
\listofchanges


\end{document}